%% file: main.tex
\pdfoutput=1

\documentclass[11pt]{article}

\usepackage{acl}

\usepackage{times}
\usepackage{latexsym}
\usepackage{booktabs}
\usepackage{booktabs}
\usepackage{amsfonts}
\usepackage{nicefrac}
\usepackage{microtype}
\usepackage{todonotes}
\usepackage{listings}
\usepackage{algorithm}
\usepackage{algpseudocode}
\usepackage{soul}
\usepackage{tabularx}
\usepackage{pifont}
\usepackage{xcolor}
\usepackage{graphicx}
\usepackage{subcaption}
\usepackage{hyperref}
\usepackage{comment}
\usepackage{tcolorbox}
\usepackage{nicematrix}
\usepackage{tabu}
\usepackage{colortbl}
\usepackage{xcolor}
\usepackage{makecell}

 \usepackage{array}
 \usepackage{threeparttable}
 \usepackage{xcolor}
 
\usepackage[normalem]{ulem}

\input{macro}

\input{dfn}

\usepackage[T1]{fontenc}

\usepackage[utf8]{inputenc}

\usepackage{microtype}

\usepackage{hyperref}
\usepackage{multirow}
\usepackage{graphicx}
\usepackage{pdfpages}
\usepackage{url}
\usepackage{xcolor}
\usepackage{epsfig}
\usepackage{adjustbox}
\usepackage{amsfonts}
\usepackage{amsmath}
\usepackage{amssymb}
\usepackage{booktabs} 
\usepackage{comment}
\usepackage{caption}
\usepackage{subcaption}
\usepackage{textcomp}
\usepackage{relsize}
\usepackage{stmaryrd}
\usepackage{bbm}
\usepackage{rotating}
\usepackage{helvet}
\usepackage{courier}
\usepackage{natbib}
\usepackage{cleveref}
\usepackage{xspace}
\usepackage{enumitem}
\usepackage{makecell}
\PassOptionsToPackage{table}{xcolor}

\title{\textsc{VerifiNER}: Verification-augmented NER via Knowledge-grounded Reasoning with Large Language Models}

\author{
    Seoyeon Kim\thanks{$^\ast$Equal contribution}~~~
    Kwangwook Seo$^\ast$~~~
    Hyungjoo Chae~~~
    Jinyoung Yeo~~~
    Dongha Lee\thanks{\ \ Corresponding author}\\
    Yonsei University\\
    \texttt{\{emseoyk,tommy2130,mapoout,jinyeo,donalee\}@yonsei.ac.kr}\\   
}

\begin{document}
\maketitle
\input{0_abs}
\input{1_intro_new}

\input{2_prelim}

\input{3_method}

\input{4_exp}

\input{5_results}

\input{2_rel}

\input{6_conc}

\input{7_limit}

\bibliography{custom}

\newpage
\appendix
\input{8_appendix}

\end{document}

%% file: macro.tex
\definecolor{mypink1}{rgb}{0.858, 0.188, 0.478}

\definecolor{olive}{rgb}{0.39, 0.875, 0.12}

\newcommand{\ie}{{\it i.e.}}%

\definecolor{yellow-green}{rgb}{0.3, 0.5, 0.0}

\definecolor{lightblue}{RGB}{224,236,247}
\definecolor{deepblue}{RGB}{9,46,107}
\definecolor{defaultcolor}{gray}{0.9}
\definecolor{green}{RGB}{0, 153, 0}

\tcbset{
  aibox/.style={
    width=\textwidth,
    top=10pt,
    colback=white,
    colframe=black,
    colbacktitle=black,
    center title,
    fonttitle=\bfseries,
  }
}

\newtcolorbox{AIbox}[2][]{aibox,title=#2,#1}

\tcbset{
  aiboxsmall/.style={
    width=0.48\textwidth,
    top=10pt,
    colback=white,
    colframe=black,
    colbacktitle=black,
    center title,
    fonttitle=\bfseries,
  }
}

\newtcolorbox{AIboxSmall}[2][]{aiboxsmall,title=#2,#1}

%% file: dfn.tex
\newcommand{\gptner}{GPT-NER\xspace}
\newcommand{\conner}{ConNER\xspace}
\newcommand{\biobert}{BioBERT\xspace}

\newcommand{\bcfive}{BC5CDR\xspace}
\newcommand{\genia}{GENIA\xspace}

\newcommand{\verifiner}{\textsc{VerifiNER}\xspace}

%% file: 0_abs.tex
\begin{abstract}

Recent approaches in domain-specific named entity recognition (NER), such as biomedical NER, have shown remarkable advances.
However, they still lack of faithfulness, producing erroneous predictions.
We assume that knowledge of entities can be useful in verifying the correctness of the predictions.
Despite the usefulness of knowledge, resolving such errors with knowledge is nontrivial, since the knowledge itself does not directly indicate the ground-truth label.
To this end, we propose \verifiner, a post-hoc verification framework that identifies errors from existing NER methods using knowledge and revises them into more faithful predictions.
Our framework leverages the reasoning abilities of large language models to adequately ground on knowledge and the contextual information in the verification process.
We validate effectiveness of \verifiner through extensive experiments on biomedical datasets. 
The results suggest that \verifiner can successfully verify errors from existing models as a model-agnostic approach.
Further analyses on out-of-domain and low-resource settings show the usefulness of \verifiner on real-world applications.\footnote{Our code and data are available at \url{https://github.com/emseoyk/VerifiNER}.}

\end{abstract}

%% file: 1_intro_new.tex
\section{Introduction}

Named entity recognition (NER) is a fundamental task in natural language processing that aims to identify entity mentions in input text and assign them to specific types~\citep{mikheev1999named, lample2016neural}.
Previous works solve NER tasks by training models on human-annotated datasets~\citep{Kim2003genia, Li2016bc5cdr, zhang-etal-2021-pdaln-other}, improving neural model architectures~\citep{Jeong2023conner}, or leveraging external knowledge~\citep{liu-etal-2019-towards-other, mengge-etal-2020-coarse-other, wang-etal-2021-improving-others}.
More recently, simply prompting large language models (LLMs) without training can also perform NER tasks~\citep{wang2023gptner, ashok2023promptner}.

Despite the promising results of these approaches, they still produce plausible but imprecise outputs. The risk to produce such errors is especially salient in domains that require expert-level knowledge such as the biomedical domain. 
Taking Figure~\ref{fig:error_case} for example, \textit{``NF-kappa B''} is incorrectly labeled as ``\texttt{RNA}'' type and \textit{``endothelia cells''} is identified with wrong span.
Erroneous prediction is a significant threat to their application in domains where high precision is required~\citep{dai2021recognising, karim2023explainable}. 


\input{figure_latex/figure1}

In this paper, we seek to minimize these errors by incorporating knowledge into the inference process. Figure~\ref{fig:error_case} suggests that knowledge can serve as a useful evidence for humans to verify that the type of ``\textit{NF-kappa B}'' is ``\texttt{protein}''. 
However, avoiding the errors during inference time is challenging for neural models, because they lack in factual evidence that can assist the models to judge the correctness of predictions. 
Furthermore, even if the models are equipped with knowledge, there exists a mismatch between knowledge and entity prediction that discourages models from properly detecting and correcting the errors. For example, the definition of ``\textit{NF-kappa B}'' as ``an ubiquitous, inducible, nuclear transcriptional activator'' does not explicitly indicate that the type is ``\texttt{protein}''. 



To resolve the aforementioned challenge, we aim to propose a post-hoc verification framework that identifies errors from existing NER methods and revises them into more faithful predictions.
We employ knowledge base (KB) to provide factual evidence to existing NER systems, and leverage reasoning and in-context learning ability of LLMs to verify entities via knowledge-grounded reasoning.
By verifying errors of the previous models in a post-hoc manner, our work exhibits remarkable performance without re-training models.

To this end, we introduce \verifiner, a novel \underline{Verifi}cation-augmented \underline{NE}R via Knowledge-grounded \underline{R}easoning. 
With the notion that the errors can be alleviated with factual knowledge and contextual cue, our framework verifies entities in terms of \textit{factuality} and \textit{contextual relevance}. 
For factuality verification, \verifiner formulates queries from each predicted entity to retrieve knowledge from KB, then reassigns its span and type based on the knowledge. 
The contextual relevance verification module employs the reasoning ability of LLMs to consider input context when selecting correct entity from candidates that are passed from previous step. 




\input{figure_latex/pie}
For effective demonstration of our framework, we conduct experiments on biomedical NER which requires domain-expert knowledge to solve the task. 
While our approach is model-agnostic, we validate that \verifiner~can be applied to both fine-tuned models~\citep{Lee2019biobert, Jeong2023conner} and a LLM-based NER method~\citep{wang2023gptner} with significant improvement. 
In addition, we evaluate our framework under out-of-distribution and low-resource settings, showing its advantages on real-world scenarios.

To the best of our knowledge, this is the first work that solves NER with a verification module that exploits reasoning ability of LLMs. The main contributions of this work are as follows:
\begin{itemize}[leftmargin=*,topsep=2pt,itemsep=2pt,parsep=0pt]
    \item We propose a novel framework for identifying and resolving errors via knowledge-grounded reasoning by utilizing knowledge and LLMs.
    \item We present \verifiner, a post-hoc 
 verification module that corrects entity prediction with respect to factuality and contextual relevance. 
    \item We demonstrate the effectiveness and generalization ability of \verifiner~through extensive experiments and analyses.
\end{itemize}


%% file: figure_latex/figure1.tex
\begin{figure}[!t]
    \centering
    \includegraphics[width=1\columnwidth]{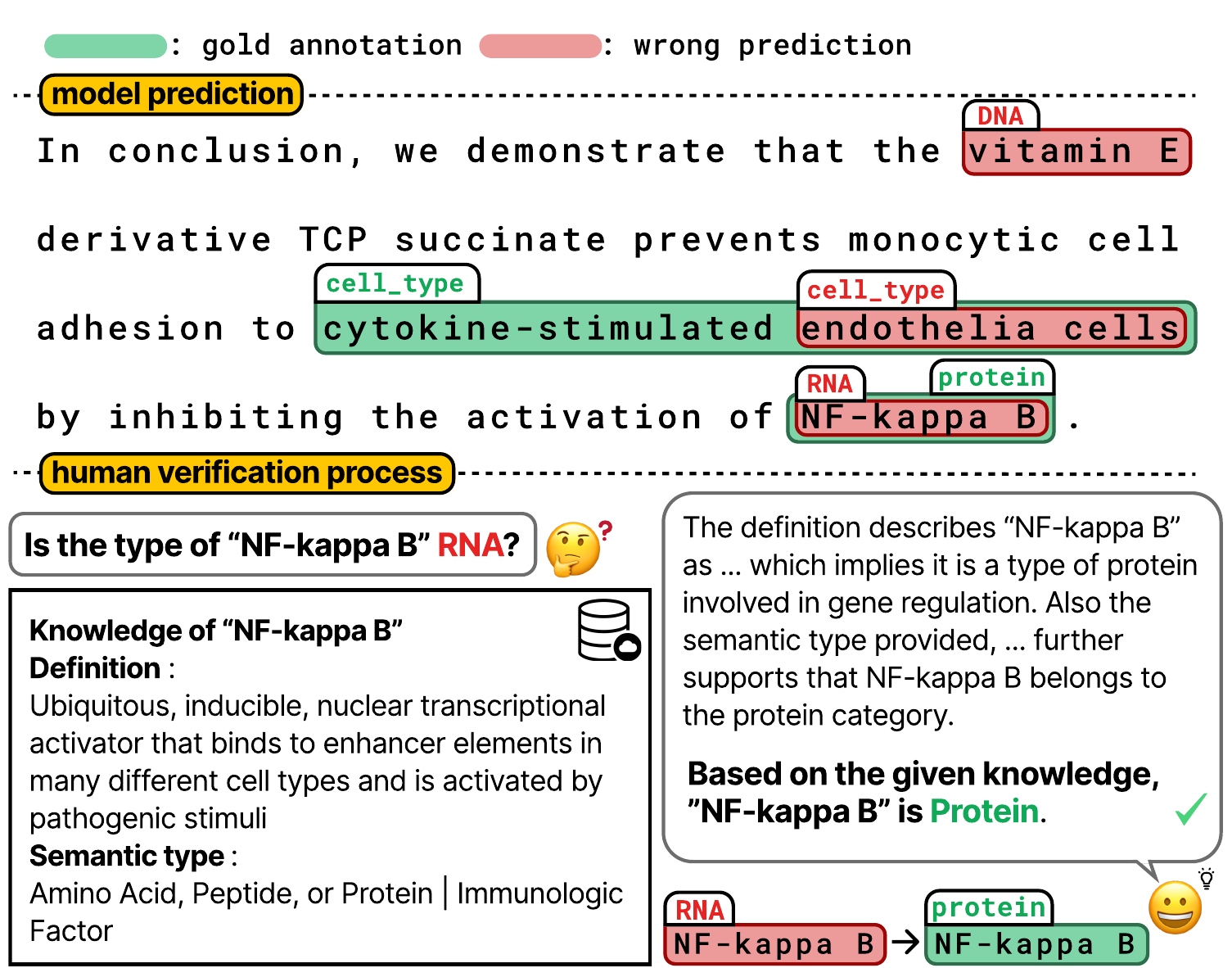}
    \caption{\textit{Top:} Erroneous prediction of ConNER model on GENIA dataset. \textit{Bottom:} Human verification process of correcting errors with knowledge from a KB. 
    From knowledge, it is evident that labeling \texttt{RNA} is incorrect.
    }
    \label{fig:error_case} 
\end{figure}

%% file: figure_latex/pie.tex
\begin{figure}[!t]
    \centering
    \includegraphics[width=1\columnwidth]{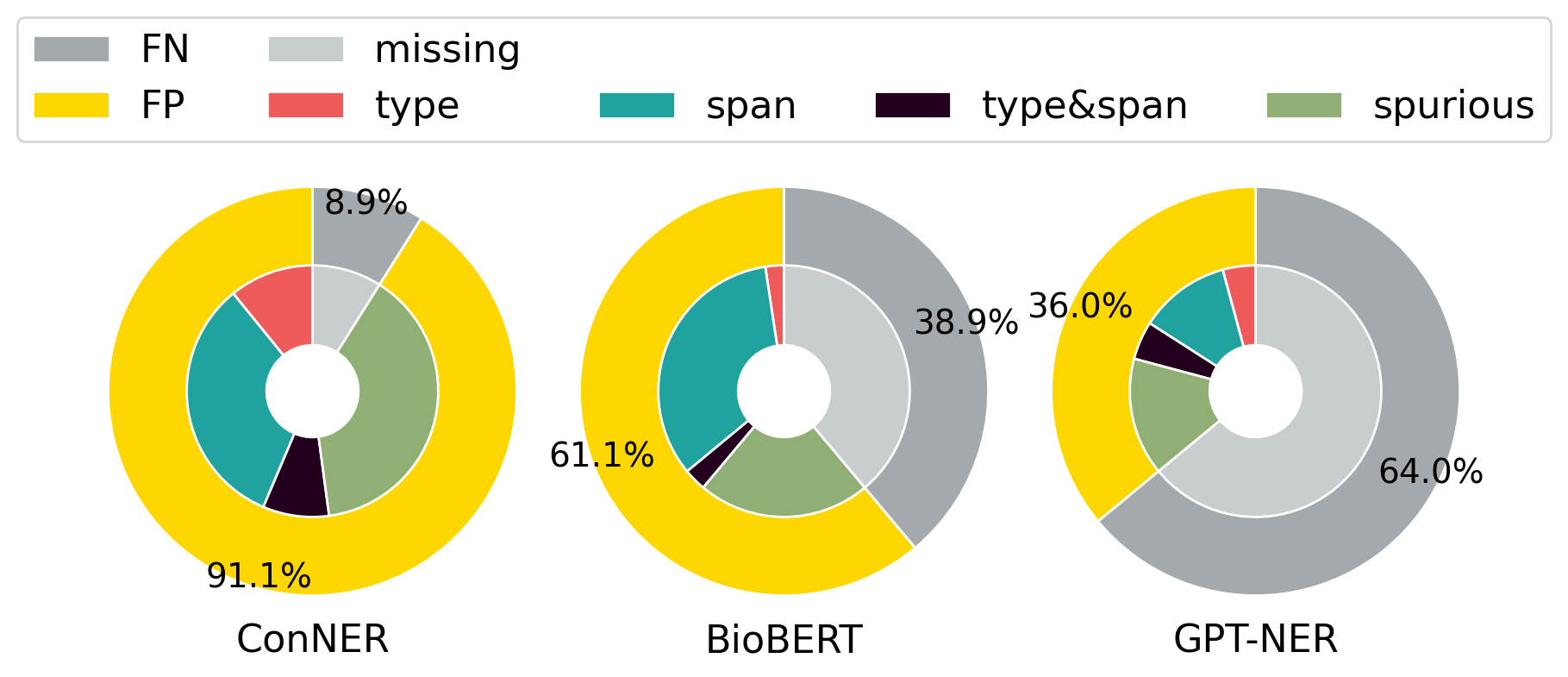}
    \caption{Ratio of error types (\%) of NER models. Detailed statistics are reported in Table~\ref{tab:pie_stat}.}
    \label{fig:error_pie} 
\end{figure} 

%% file: 2_prelim.tex

\input{figure_latex/token_diff}
\input{tables/error_types}

\section{Preliminary Analysis}
\label{sec:prelim}


In this section, we first define the common error types of NER we aim to resolve. 
Then, we provide further analysis on error cases in existing models to gain insight on how to rectify the errors.

\paragraph{Error Types in NER.}
For evaluating NER model predictions against the gold annotations, error cases can be categorized as either false positives (FP) or false negatives (FN) based on whether the model incorrectly identifies entities or fails to recognize the gold entities that is present.
Within the FP, errors can occur from mismatches of type or span.

Depending on the combinations of these mismatches, errors are classified into more fine-grained error types, including Type, Span, Type\&Span and Spurious errors.
We provide definitions and examples of each error type in Table~\ref{tab:error_def}.

\paragraph{Error Analysis on NER Models.}
We conduct an analysis on existing errors using two fine-tuned models, \ie{}, ConNER~\cite{Jeong2023conner}, BioBERT~\cite{Lee2019biobert},  and one prompting-based LLM, \ie{}, GPT-NER~\cite{wang2023gptner}. 
Figure~\ref{fig:error_pie} shows the ratio of different error types for each model.
In case of the fine-tuned models, false positive errors take the majority of the total error types. 
For prompting-based LLM, FP cases take more than one third. 
As low precision in domain-specific NER is a crucial problem ~\citep{dai2021recognising, karim2023explainable} and FP constitute a significant proportion across all three models, we focus on correcting the FP cases from the initial predictions. 
By doing so, we can boost the NER performance regardless of the model that comes beforehand.

To gain clues on how to resolve the errors, we examine where existing models fall short to make precise predictions.
Among the FP cases, we take a closer look on errors (\ie{}, Type, Span, and T\&S) where the models' predictions partially overlap with the gold annotation, but either the type or span is incorrect.
In Figure~\ref{fig:error_pie}, almost 60\% of FP cases are partially overlapping predictions for all three models. 
For span error, in Figure~\ref{fig:token_diff}, more than 80\% of predicted entities show difference in length with gold annotation within a margin of two tokens. 
This implies that NER models resort to plausible yet incorrect predictions with a small margin.
Thus, instead of completely discarding the wrong entity predictions, we use them as queries to initialize our verification process in a post-hoc manner.

%% file: figure_latex/token_diff.tex
\begin{figure}[!t]
    \centering
    \includegraphics[width=1\columnwidth]{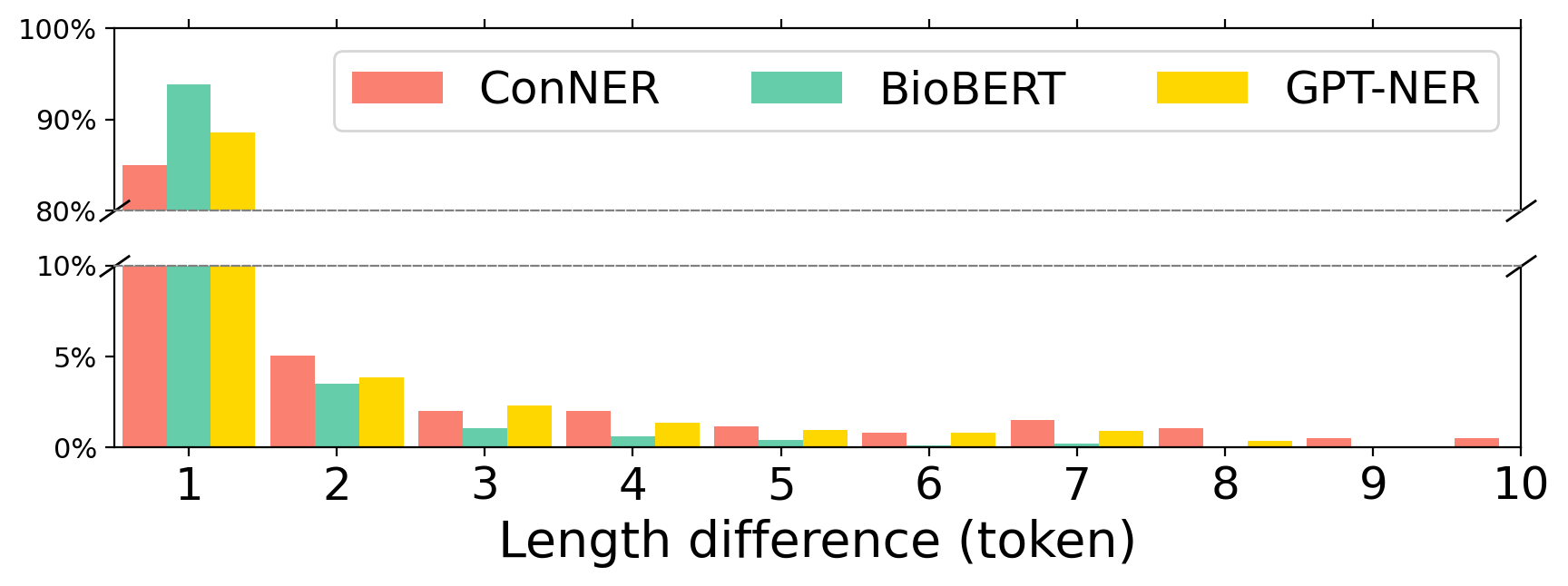}
    \caption{Distribution of token-level deviation between predicted and gold spans for each model. The x-axis represents the difference in token length between entities where the predicted span and gold span overlaps.}
    \label{fig:token_diff} 
\end{figure} 

%% file: tables/error_types.tex
\newcolumntype{C}{>{\centering\arraybackslash}p{0.1\columnwidth}}
\newcolumntype{R}{>{\raggedleft\arraybackslash}p{0.1\columnwidth}}
\newcolumntype{L}{>{\raggedright\arraybackslash}p{0.35\textwidth}}
\newcolumntype{F}{>{\raggedright\arraybackslash}p{0.55\textwidth}}
\newcommand*{\mline}[1]{%
\begingroup
    \renewcommand*{\arraystretch}{1.1}%
   \begin{tabular}[c]{@{}>{\raggedright\arraybackslash}p{2cm}@{}}#1\end{tabular}%
  \endgroup
}

\begin{table*}[!t]
\centering
\small
\renewcommand{\tabcolsep}{1mm}
\resizebox{.99\textwidth}{!}{%
\begin{tabular}{ccLl}
\toprule
\multicolumn{4}{l}{
\textbf{Gold annotation:} ``… cell adhesion to cytokine-stimulated \textcolor{green}{{[endothelial cells]}{\textsubscript{\texttt{cell\_type}}}} by ..''
} \\
\toprule
\textbf{Category}
& \textbf{Error type}& \multicolumn{1}{c}{\textbf{Definition}} & \multicolumn{1}{c}{\textbf{Example}} \\
\midrule
\renewcommand{\arraystretch}{1.2}
\multirow{7}{*}{FP} & Type & Wrong type is assigned to an entity & "… cell adhesion to cytokine-stimulated \textcolor{green}{[endothelial cells]} \textcolor{red}{\textsubscript{\texttt{cell\_line}}} by .." \\
\cmidrule{2-4}
& Span & Predicted span partially overlaps with ground truth, but incorrect & "… cell adhesion to \textcolor{red}{[cytokine-stimulated endothelial cells]} \textcolor{green}{\textsubscript{\texttt{cell\_type}}} by .." \\
\cmidrule{2-4}
& T\&S & Both type and span are predicted incorrectly & "… cell adhesion to \textcolor{red}{[cytokine-stimulated endothelial cells] \textsubscript{\texttt{cell\_line}}} by .." \\
\cmidrule{2-4}
& Spurious & A completely incorrect entity is predicted where gold annotation does not exist & "… \textcolor{red}{[cell adhesion] \textsubscript{\texttt{cell\_line}}} to cytokine-stimulated endothelial cells by .." \\
\midrule
\multirow{2}{*}{FN} & Missing & A gold annotation for entity exists but not predicted by a model & "… cell adhesion to cytokine-stimulated endothelial cells by .." \\
\bottomrule
\end{tabular}
} 
\caption{\textit{Top:} Ground truth span and type of ``\textit{endothelial cells}''. 
\textit{Bottom:} Definition and example per error type. T\&S stands for Type\&Span. \textcolor{green}{Green} denotes ground truth, and \textcolor{red}{red} denotes error in span and/or type.}
\label{tab:error_def}
\end{table*}

%% file: 3_method.tex
\input{figure_latex/figure2}

\section{VerifiNER}
In this section, we propose \verifiner, a framework that verifies errors from NER models with external knowledge and the reasoning ability of LLMs.
In this paper, we define the term \textit{verification} as a process that also includes revision. Motivated by the observations in Section~\ref{sec:prelim}, we design our framework to particularly focus on enhancing precision; that is, we first identify errors in predicted entities and then correct them into accurate outputs.
The overall framework is illustrated in Figure~\ref{fig:overview}.

Formally, given an input sequence $\mathcal{X}=\{x_1, x_2, \cdots, x_n\}$ of $n$ tokens and a predefined type set $\mathcal{T}$, 
our goal is to produce a revised entity prediction $\bar{e}$ by verifying an entity $e =(s, t)$ that is originally predicted by an off-the-shelf NER model, where $s=[x_\texttt{beg}: x_\texttt{end}]$ and $t \ (\in \mathcal{T})$ represent its span and type, respectively. 
In this sense, the verification process targets both the span and type of the entity.
Note that span verification must precede type verification, because predicting the type $t$ of an entity $e$ depends on the semantics of $e$ from its span $s$.
For example, the type of ``\textit{PEBP2}'' is ``\texttt{protein}'', but a longer span, ``\textit{PEBP2 site}'', is classified as ``\texttt{DNA}''; 
thus, we first identify spans and then proceed to determine types.

Specifically, we first verify the factuality of the span and type (Sections~\ref{sec:span} and~\ref{sec:type}), then verify whether the factually approved entity is relevant to the input context (Section~\ref{sec:context}) for precise revision.


\subsection{Span Factuality Verification}
\label{sec:span}
In this step, we rectify span error through collecting candidate spans from the predicted entity and verifying them by leveraging a KB. 
Based on the observation in Section~\ref{sec:prelim}, we assume that the gold entity is likely to be adjacent to the predicted entity.
Therefore, we  expand the range of candidate spans around the predicted entity span $s$ in both directions to increase the likelihood of the gold entity within our candidate set. 

We collect a set of candidate spans $\Tilde{\mathcal{S}}$ by extending the left and right offsets of a span $s$ with hyperparameter $\alpha$, then enumerate sub-sequences within the offsets $[x_{\texttt{beg}-\alpha}:x_{\texttt{end}+\alpha}]$, so that each candidate partially or completely overlaps with the span $s$ of the predicted entity $e$. 
As shown in Figure~\ref{fig:overview} (a), \{``\textit{human mononuclear}'', ``\textit{human}'', ``\textit{from human}'', \textit{...}, ``\textit{mononuclear}'', ``\textit{mononuclear leukocytes}'', ``\textit{human mononuclear leukocytes}''\} can be extracted from the predicted entity ``\textit{human mononuclear}''. 

Then, we search the KB to prune only factually valid candidates.\footnote{We assume a domain-specific NER scenario where all gold entities exist entirely within the KB. More explanation is in Appendix~\ref{sec:appendix_implemetation_details}. } 
Using each candidate span $\Tilde{s}\ (\in \Tilde{\mathcal{S}})$ as a query, we check whether knowledge for each candidate exists in the KB. 
If a candidate is found in the search, we consider its factuality is verified and collect its associated knowledge $k$. 
On the other hand, if the candidate is not defined in the KB, we assume it is a noisy candidate and do not consider it further in the remaining process.
For example, in Figure~\ref{fig:overview} (a), the search result for ``\textit{human mononuclear}'' is not found, and this candidate is excluded from the next step. In consequence, we obtain a pruned set of candidate spans $\Tilde{\mathcal{S}}$.

\input{figure_latex/evidence}
\subsection{Type Factuality Verification}
\label{sec:type}
As we collect the set of candidate spans $\Tilde{s}$ and their associated knowledge $k$, we  proceed to re-assign types to $\Tilde{s}$ grounded on the retrieved knowledge. 
While knowledge serves as a reliable source for verifying candidates, directly applying it for verification is challenging. It is due to the fact that knowledge often lacks explicit indications regarding whether a candidate is correctly labeled. 
In Figure~\ref{fig:evidence}, the knowledge of ``\textit{mononuclear leukocytes}'', \ie{}, ``a white blood cell ...'' and ``Quantitative Concept'' do not exactly match with ``\texttt{cell\_type}''. 

To this end, we leverage the reasoning ability of LLMs to project knowledge into predefined types. 
We accomplish this by generating evidence $k^\prime$ to assist in assigning types grounded on knowledge. 
Specifically, we rationalize the source knowledge $k$ into verbalized form to generate $k^{\prime}$. 
Then, we provide {($\Tilde{s}$, $k^{\prime}$)} with the predefined label set $\mathcal{T}$ and prompt the LLM to re-assign type $\Tilde{t}$ based on the evidence $k^\prime$. 
An example of knowledge-grounded evidence is provided in Figure~\ref{fig:evidence}.
Based on provided definition and semantic type of the candidate ``\textit{mononuclear leukocytes}'', LLM generates a knowledge-grounded evidence and assigns type to the candidate. 

If the knowledge is irrelevant to the domain, the LLM will assign $\Tilde{t}$ as \texttt{NONE}. 
Consequently, each entity will have a set of factuality-verified candidates $\Tilde{e}\in \Tilde{\mathcal{E}}$, where each candidate is $\Tilde{e} = (\Tilde{s}, \Tilde{t})$. 

\input{figure_latex/Contextual_Relevance}

\subsection{Contextual Relevance Verification}
\label{sec:context}

Lastly, we select a final candidate entity from $\Tilde{\mathcal{E}}$ based on contextual relevance for revising the prediction $e$ to $\bar{e}$.
A candidate $\Tilde{e}$ can be a valid entity $\bar{e}$ if (1) it is semantically relevant to the input context $\mathcal{X}$, and (2) its knowledge-grounded evidence $k^{\prime}$ is aligned well with the context compared to other candidates. To determine the final candidate grounded on knowledge and context, we employ in-context learning ability of LLMs.

When verifying contextual relevance of the candidate, both local and global contextual information should be considered. However, reasoning only once may lead to insufficient attention on limited contexts.
Therefore, we sample multiple reasoning paths to gather answers that reflect various aspects of the context. 
This process resembles the human annotation process of gathering various opinions from multiple annotators and converging them into a single consistent answer through discussion. 
To facilitate this process, we employ self-consistency~\citep{wang2023selfconsistency} and use consistency voting to select the candidate that is most suitable for the context. 
As demonstrated in Figure~\ref{fig:reasoning_path}, LLMs generate reasoning paths that are properly grounded on knowledge. 


Given the input context $\mathcal{X}$, candidates $\Tilde{\mathcal{E}}$, and evidence for each candidate $k^{\prime}$, we prompt the LLMs to sample $N$ reasoning paths where each path selects a single candidate that has the most faithful evidence and is relevant to $\mathcal{X}$.
We conduct majority voting over the collected $N$ answers and select the answer that receives the most votes. 
Finally, the prediction is revised as $\bar{e}=(\bar{s}, \bar{t})$. 
The prompts used in our framework are provided in Appendix~\ref{sec:prompts}.

%% file: figure_latex/figure2.tex
\begin{figure*}[!ht]
    \centering
    \includegraphics[width=1\textwidth]{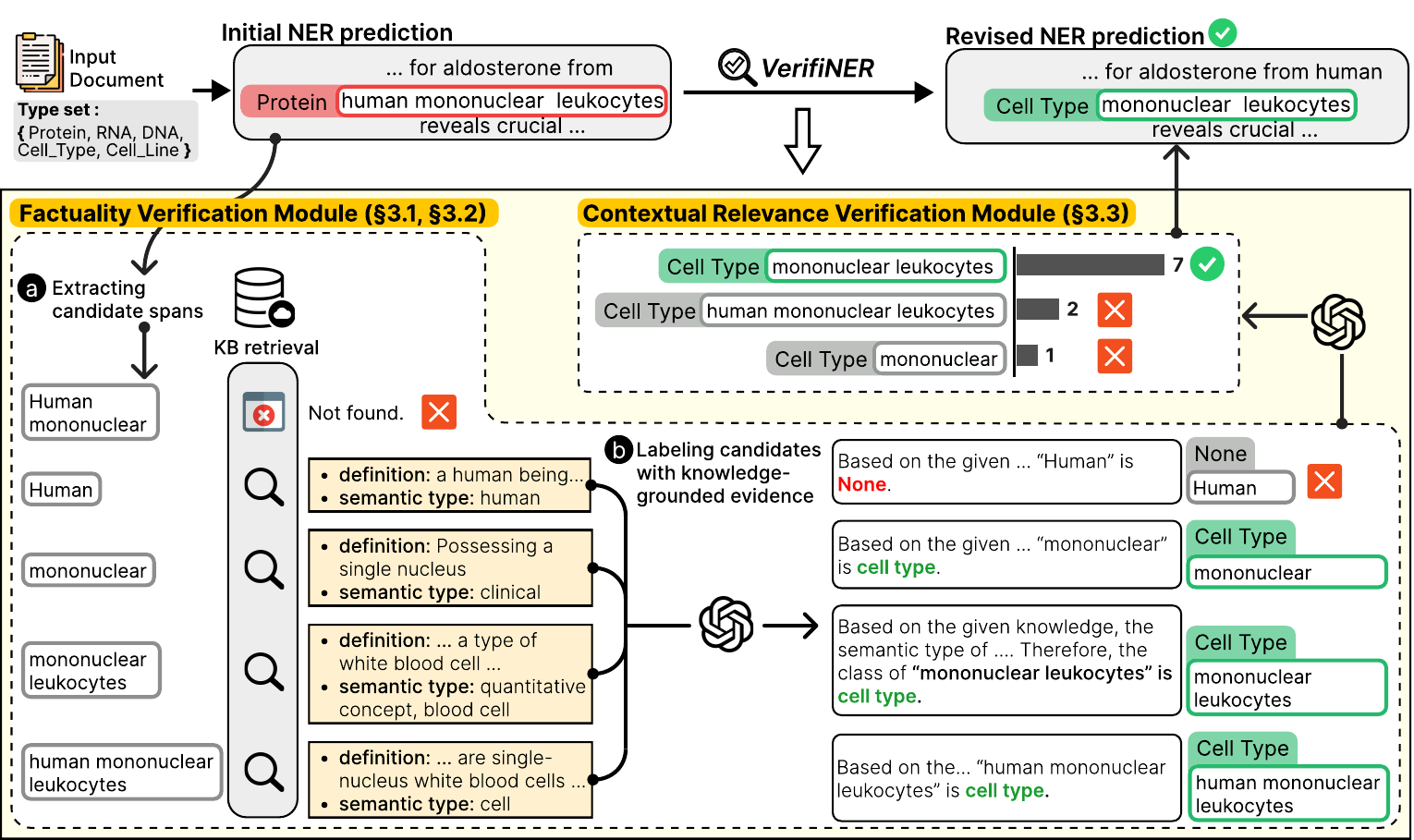}
    \caption{Overview of \verifiner framework. 
    (1) Using entity prediction by existing models, we (a) extract candidate spans to retrieve knowledge from KB and verify \textit{factuality} of span accordingly. 
    Then (b) using retrieved knowledge, we verify \textit{factuality} of type by generating knowledge-grounded evidence. 
    (2) Lastly, we take consistency voting to select a candidate that is the most \textit{contextually relevant}, with the help of the reasoning ability of LLMs. 
    }
    \label{fig:overview} ~
\end{figure*} 

%% file: figure_latex/evidence.tex

    
   
    

\begin{figure}[!t]
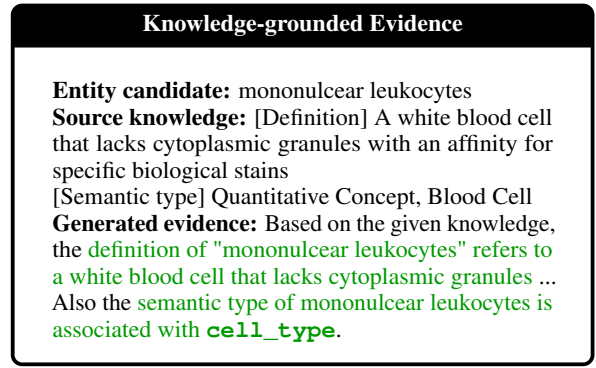

\tiny
    \begin{AIboxSmall}{\footnotesize Knowledge-grounded Evidence}
    \footnotesize
    \textbf{Entity candidate:} mononulcear leukocytes

    \textbf{Source knowledge:} 
    [Definition] A white blood cell that lacks cytoplasmic granules with an affinity for specific biological stains 
    
    [Semantic type] Quantitative Concept, Blood Cell
   
    \textbf{Generated evidence:} Based on the given knowledge, the \textcolor{green}{definition of "mononulcear leukocytes" refers to a white blood cell that lacks cytoplasmic granules} ... Also the \textcolor{green}{semantic type of mononulcear leukocytes is associated with \texttt{\textbf{cell\_type}}}. 
    
    \end{AIboxSmall}
    \caption{Example of generated evidence used for type factuality verification.}
    \label{fig:evidence}
\end{figure}

%% file: figure_latex/Contextual_Relevance.tex
\begin{figure}[!t]
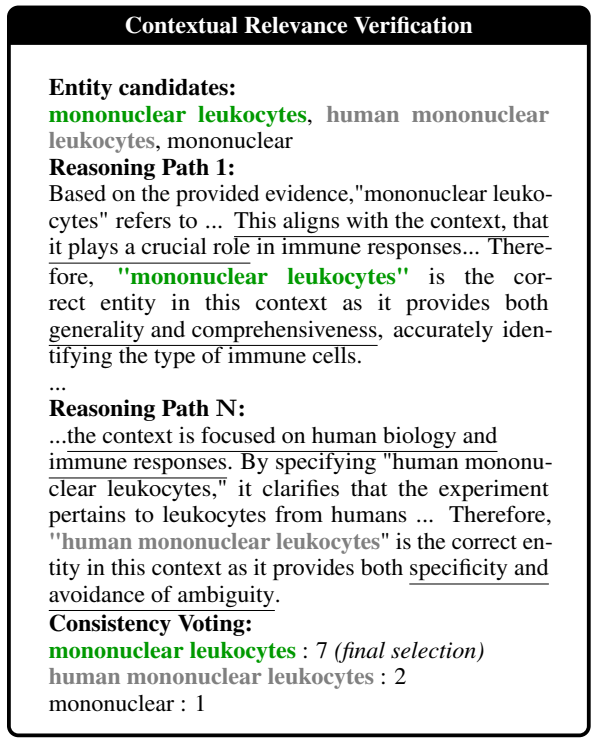

\tiny
    \begin{AIboxSmall}{\footnotesize Contextual Relevance Verification}
    \footnotesize
    \textbf{Entity candidates:} 
    
    \textcolor{green}{\textbf{mononuclear leukocytes}}, \textcolor{gray}{\textbf{human mononuclear leukocytes}}, mononuclear
    
    \textbf{Reasoning Path 1:}
    
    Based on the provided evidence,"mononuclear leukocytes" refers to ... \underline{This aligns with the context, that} \underline{it plays a crucial role} in immune responses...
    Therefore, \textcolor{green}{\textbf{"mononuclear leukocytes"}} is the correct entity in this context as it provides both \underline{generality and comprehensiveness}, accurately identifying the type of immune cells. 
    
    ...
    
    \textbf{Reasoning Path $\bold{N}$:}

    ...\underline{the context is focused on human biology and} \underline{immune responses}. 
    By specifying "human mononuclear leukocytes," it clarifies that the experiment pertains to leukocytes from humans ... 
    Therefore, \textcolor{gray}{\textbf{"human mononuclear leukocytes}}" is the correct entity in this context as it provides both \underline{specificity and} \underline{avoidance of ambiguity}.

    \textbf{Consistency Voting:}

    \textcolor{green}{\textbf{mononuclear leukocytes}} : 7 \textit{(final selection)}
    
    \textcolor{gray}{\textbf{human mononuclear leukocytes}} : 2
    
    mononuclear : 1

    \end{AIboxSmall}

    \caption{Example of generated diverse reasoning paths for contextual relevance verification.}

    \label{fig:reasoning_path}
\end{figure}

%% file: 4_exp.tex
\input{tables/main_table}

\section{Experiments}
In this section, we design our experiments to answer the following research questions:
\begin{itemize}[leftmargin=*,topsep=2pt,itemsep=2pt,parsep=0pt]
\item \textbf{RQ1}: Can \verifiner~ faithfully identify and revise errors? 
\item \textbf{RQ2}: Can \verifiner induce generalizability to fine-tuned models in various test distributions?
\item \textbf{RQ3}: Can \verifiner effectively mitigate the low-resource challenge in domain-specific NER? 
\end{itemize}


\subsection{Experimental Settings}

We evaluate our framework on two biomedical datasets, each with different set of predefined types:
BC5CDR~\citep{Li2016bc5cdr} is annotated with \texttt{Chemical} and \texttt{Disease}, while GENIA~\citep{Kim2003genia} includes five types, including \texttt{cell\_line}, \texttt{cell\_type}, \texttt{DNA}, \texttt{RNA}, and \texttt{protein}. 
We use the Unified Medical Language System (UMLS)~\citep{Bodenreider2004} as a KB, which is a database containing over two millions of biomedical terminologies annotated with their definitions, semantic types, and lexical relationships. 
We employ ChatGPT as an LLM to implement our framework~\citep{openai2023chatgpt}.\footnote{We use \texttt{gpt-3.5-turbo-1106} model among the various versions of ChatGPT.}
Following~\citet{Jeong2023conner}, we measure the performance utilizing entity-level Precision (P), Recall (R), and F1.

\subsection{NER Models}
\label{sec:ner_models}
To validate the effectiveness of \verifiner~as a model-agnostic approach, we employ two groups of NER models as a test-bed for our evaluation:
For {fine-tuned models}, we use \textbf{\conner}~\citep{Jeong2023conner} and \textbf{\biobert}~\citep{Lee2019biobert}, trained for the NER task on each dataset. For {prompting-based LLMs}, we consider \textbf{\gptner}~\citep{wang2023gptner}, which predicts entity spans of interest and their types in a generative manner via few-shot prompting.


\subsection{Baseline Methods}
We compare \verifiner~with other methods that revise the initial prediction in a post-hoc manner.
(1) \textbf{Manual Mapping}: Naive use of external knowledge to verify predictions. 
We reassign entity types by manually mapping semantic types found in the KB to the predefined labels.
Assuming that entities with the same semantic type will have the same label accordingly, the purpose of Manual Mapping is to solely rely on the KB and not the reasoning function of LLM. 
To construct a semantic type-label map, we use entities from the train set to retrieve the semantic types from KB and manually align them with the label. For example, if the KB retrieves [``\texttt{Organic Chemical}'', ``\texttt{Pharmacologic Substance}''] as semantic types of  ``\texttt{cyproterone acetate}'' and its label is ``\texttt{Chemical}'', then we align the semantic type ``\texttt{Pharmacologic Substance}'' to the label ``\texttt{Chemical}''. 
The statistics of semantic types assigned to each label are in Appendix~\ref{sec:appendix_implemetation_details}.
(2) \textbf{LLM-revision}: Simple revision using an LLM, where the model re-examines the input context and generates revised context based on marked predicted entities.
(3) \textbf{LLM-revision w/ CoT}: We incorporate zero-shot CoT~\citep{NEURIPS2022_zeroshotcot} in addition to LLM-revision.  We provide prompts for LLM-revision and w/CoT in Appendix~\ref{sec:prompts}.




%% file: tables/main_table.tex
\newcolumntype{W}{>{\centering\arraybackslash}m{0.06\linewidth}}

\begin{table*}[!ht]
\small
\centering
\renewcommand{\tabcolsep}{1.5mm}
\begin{tabular}{lWWWWWWWW}
\toprule
  \multirow{2.5}{*}{\textbf{Methods}} &
  \multicolumn{4}{c}{\textbf{\genia}} &
  \multicolumn{4}{c}{\textbf{\bcfive}} \\ 
  \cmidrule(lr){2-5} \cmidrule(lr){6-9}

\textbf{} & 
  P &
  R &
  F1 &
  $\Delta$F1 &
  P &
  R &
  F1 &
  $\Delta$F1
  \\ 
\midrule
\gptner~\citep{wang2023gptner} &   56.44 & 42.15 & 48.26 & -& 79.84 & \textbf{47.48} & 59.55 & -\\ 
\hspace{0.1cm}+ Manual Mapping &            37.53 & 32.65 & 34.93 & -13.33 & 51.82 & 36.98 & 43.16 & -16.39\\
\hspace{0.1cm}+ LLM-revision &              52.97 & \textbf{46.77} & 49.68 & +1.42 & 77.21 & 44.53 & 56.48 & -3.07 \\
\hspace{0.1cm}+ LLM-revision w/ CoT &       53.57 & 44.54 & 48.64 & +0.38 & 76.49 & 44.91 & 56.59 & -2.96\\
\rowcolor{defaultcolor}\hspace{0.1cm}\textbf{+ \verifiner~(Ours)} & 
                                            \textbf{72.37} & 44.95 & \textbf{55.46} & \textbf{+7.20} &\textbf{91.01} & 46.92 & \textbf{61.92} & \textbf{+2.37}\\
\midrule

\conner~\citep{Jeong2023conner}&    74.13 & \textbf{96.69} & 83.92 & - & 84.90 & \textbf{96.47} & 90.32 & -\\
\hspace{0.1cm}+ Manual Mapping &            43.62 & 94.50 & 59.69 &  -24.23 & 53.98 & 94.52 & 68.71 & -21.61\\
\hspace{0.1cm}+ LLM-revision &              63.64 & 86.64 & 73.38 & -10.54  & 80.35 & 93.07 & 86.25 & -4.07\\
\hspace{0.1cm}+ LLM-revision w/ CoT &       64.85 & 86.92 & 74.28 & -9.64 & 78.14 & 92.99 & 84.92 & -5.40\\
\rowcolor{defaultcolor}\hspace{0.1cm}\textbf{+ \verifiner~(Ours)} & 
                                            \textbf{79.07} & 91.82 & \textbf{84.97} & \textbf{+1.05} & \textbf{94.77} & 91.61 & \textbf{93.16}& \textbf{+2.84}\\
 \midrule

\biobert~\citep{Lee2019biobert} &   54.51 & 65.30 & 59.42 & - & 79.93 & \textbf{95.98} & 87.22 & -\\
\hspace{0.1cm}+ Manual Mapping &            30.57 & 24.39 & 27.14 & -32.28 & 38.65 & 65.78 & 48.69 & -38.53 \\
\hspace{0.1cm}+ LLM-revision &              52.63 & 65.01 & 58.17 & -1.25 & 60.79 & 77.74 & 68.23 & -18.99\\
\hspace{0.1cm}+ LLM-revision w/ CoT &       52.21 & 63.49 & 57.30 & -2.12 & 59.43 & 78.66 & 67.71& -19.51 \\
\rowcolor{defaultcolor}\hspace{0.1cm}\textbf{+ \verifiner~(Ours)} & 
                                            \textbf{77.45} & \textbf{67.75} & \textbf{72.31} & \textbf{+12.89} & \textbf{94.02} & 91.17 & \textbf{92.57} & \textbf{+5.35}\\
\bottomrule

\end{tabular}
\caption{Results of \verifiner on \genia and \bcfive compared to baselines. The performance is evaluated on test set based on the entity-level exact matching. $\Delta$F1 indicates the improvement on F1 from the initial models. }
\label{tab:main_results}
\end{table*}

%% file: 5_results.tex

\section{Results}
\subsection{Effectiveness of \verifiner (RQ1)}
\label{ssec:baseline}



\paragraph{Comparison with Baselines.}
From the results in Table~\ref{tab:main_results}, we have the following observations: 
(1) For all NER models, \verifiner consistently achieves significant improvements over initial predictions on both datasets.
This demonstrates the effectiveness of our model-agnostic verification method.
(2)\verifiner~ also outperforms other revision baselines by notable margins. 
When comparing LLM-revision and w/ CoT to ours, we find that relying solely on the internal knowledge of LLMs degrades their performance, necessitating a reliable external knowledge source to faithfully verify entity predictions.
(3) While the incorporation of reliable knowledge is essential for verification, the performance drop in Manual Mapping additionally highlights the need for an intermediate reasoning process to bridge the gap between the retrieved knowledge and model predictions. 
(4) As \verifiner aims to precisely correct entity predictions, an inevitable decrease in recall is expected. However we notice that \verifiner achieves considerable increase in precision across all NER models without a severe drop in recall.
Across all datasets and models, the average increase in precision is 20.03\%, while recall decreases by only 1.09\%.
This indicates that our approach can faithfully correct errors without imposing a significant trade-off between precision and recall.
We provide further case study in Appendix~\ref{sec:case_study}.



\input{tables/ablation}
\paragraph{Ablation Study.}
To validate the effectiveness of each component in our framework, we conduct ablation studies as follows: 
excluding consistency voting ({w/o CV}), omitting knowledge-grounded evidence generation and consistency voting ({w/o evidence, CV}), generating evidence based on internal knowledge of LLMs ({w/o KB}).
In Table~\ref{tab:ablation}, we have the following observations for each ablated approach: 
(1) Verifying contextual relevance with multiple reasoning paths is helpful for enhanced accuracy, suggesting that diverse aspects within the context should be considered to select the consistent answer. 
(2) Knowledge-grounded evidence largely affects the final NER accuracy, and this implies that it is important to bridge the gap between the collected knowledge and the target application (or task).  
(3) Parametric knowledge from the LLM still lacks in domain expertise, supporting the necessity of employing external knowledge in our framework.

\paragraph{Analysis on Error Correction.}
To have a better understanding of how \verifiner works, we provide an in-depth analysis on error correction rate compared to other revision baselines.
Figure~\ref{fig:error_drop} shows that \verifiner rectifies errors more faithfully than the baselines across all error types. 
Notably, 52\% and 78\% of errors are corrected in total for \genia and \bcfive, respectively.
Upon examining the results for each error type, \verifiner successfully corrects the majority of type errors, demonstrating its effective execution of knowledge-grounded re-typing.
For span error, other revision baselines show significantly lower correction rates, while \verifiner successfully rectifies it twice as much. 
When comparing the results across datasets, \verifiner tends to struggle more on \genia compared to \bcfive, due to more fine-grained type set of \genia. 
However, even on \genia, \verifiner manages to correct over half of the errors in most cases while other revision baselines tend to remain at less than one-third.
Additionally, in the case of \bcfive, it achieves correction rates close to 90\% for spurious errors, highlighting its capability to properly filter out invalid entities.
\input{figure_latex/error_drop}

\subsection{Generalizability of \verifiner (RQ2)}
\label{ssec:unseen}
Although fine-tuned NER models show promising accuracy in in-distribution settings, \ie{}, the train and test data are sampled from the same dataset (or distribution), they can hardly be applied to unseen labels or shifted distribution, lacking robustness in out-of-distribution (OOD) settings.
Thus, we investigate if our \verifiner framework can also prove effectiveness in OOD scenarios, where fine-tuned models are trained on the datasets that are distinct from the test distribution.


\paragraph{Unseen Distribution.}
We apply \verifiner to the NER models fine-tuned on a source dataset, and evaluate it on a target dataset whose labels are not seen during the training (\ie{}, train on GENIA and infer on BC5CDR, and vice versa). 
Intuitively, the fine-tuned models themselves cannot be evaluated in the cross-dataset setting due to the different entity type sets between the source and target datasets.
Thus, we denote them as not applicable (N/A) in Table~\ref{tab:unseen_only}.
On the contrary, the fine-tuned models equipped with \verifiner perform much better than the prompting-based LLM (GPT-NER), and even achieve comparable performance to baselines trained on the target dataset. 
This suggests that \verifiner can heighten the performance of fine-tuned NER models on unseen datasets, where the training dataset is not accessible.
\input{tables/unseen_only}

\input{tables/shift_only}

\input{figure_latex/low_resource}
\paragraph{Shifted Distribution.}
Additionally, we evaluate the performance of \verifiner
in scenarios where training and test datasets have different distributions but their entity type sets are identical. 
Specifically, we split the original train data ${train}$ into subsets ${train}'_1, {train}'_2 \subset {train}$, making their shifted type distributions distinct from the distribution of target, \ie{}, $\mathcal{D}_{train'_1},\mathcal{D}_{train'_2}\neq\mathcal{D}_{test}(\approx\mathcal{D}_{train})$, but their type sets are the same, \ie{}, $\mathcal{T}_{train'_1},\mathcal{T}_{train'_2} = \mathcal{T}_{test}$.
Then, using each subset as training data, we fine-tune NER models (\ie{}, ConNER and BioBERT) to perform inference on $test$.
Please refer to Appendix \ref{sec:appendix} for experimental details and results of other subsets.
In Table~\ref{tab:unseen_shift}, \verifiner brings a significant improvement on both models.
For both \biobert and ConNER, our approach is even comparable to the in-distribution setting that is trained on the target distribution.
In particular, \verifiner even outperforms \biobert trained on the target dataset of \genia by 14.59\% on F1 score.
This highlights \verifiner's ability to generalize without overfitting to the source distribution, as it verifies predictions based on reliable external knowledge sources.

\subsection{Robustness of \verifiner in Low-resource Scenarios (RQ3)}
\label{ssec:low}
To assess how our framework can elevate robustness of NER models in low-resource scenarios, we investigate the performance changes of two fine-tuned models and those combined with \verifiner, while varying the scale of training data for fine-tuning.
For this, we generate training datasets of varying sizes, by randomly sampling examples from the original dataset across a range from 5\% to 100\%.
Figure~\ref{fig:low_resource} illustrates that post-hoc verification of \verifiner leads to notable enhancements in the performance of both \biobert and \conner across all low-resource scenarios examined in both datasets. 
Specifically, the gap between the fine-tuned models and those with \verifiner becomes large as the number of training examples decreases.
Furthermore, \verifiner consistently achieves high precision irrespective of the number of training examples.
In conclusion, \verifiner is robust in diverse low-resource scenarios within the biomedical domain, where the scarcity of available datasets has remained as a long-standing challenge.

%% file: tables/ablation.tex
\begin{table}[!t]
\centering
\resizebox{0.99\columnwidth}{!}{
\begin{tabular}{llccc}
\toprule
& \textbf{Methods}&
  P &
  R & 
  F1 \\ 
\toprule
\multirow{4}{*}{\rotatebox[origin=c]{90}{GPT-NER}} & \hspace{0.1cm}+{\verifiner} & \textbf{72.37} & \textbf{44.95} & \textbf{55.46} \\ 
&\hspace{0.3cm} w/o {\small Consistency Voting (CV)} & 71.57 & 42.95 & 53.68 \\
&\hspace{0.3cm} w/o evidence, CV & 43.49 & 34.32 & 38.36 \\
&\hspace{0.3cm} w/o KB & 46.05 & 32.93 & 38.40 \\

\midrule

\multirow{4}{*}{\rotatebox[origin=c]{90}{ConNER}} & \hspace{0.1cm}+{\verifiner} & \textbf{79.07} & \textbf{91.82} & \textbf{84.97} \\ 

&\hspace{0.3cm} w/o {\small Consistency Voting (CV)} & 77.63 & 86.14 & 81.66 \\
&\hspace{0.3cm} w/o evidence, CV & 52.70 & 91.09 & 66.77 \\
&\hspace{0.3cm} w/o KB & 50.82 & 82.42 & 62.87 \\

\bottomrule

\end{tabular}
}
\caption{Ablation results on \genia. 
}
\label{tab:ablation}
\end{table}

%% file: figure_latex/error_drop.tex
\begin{figure}[!t]
    \centering
    \includegraphics[width=1\columnwidth]{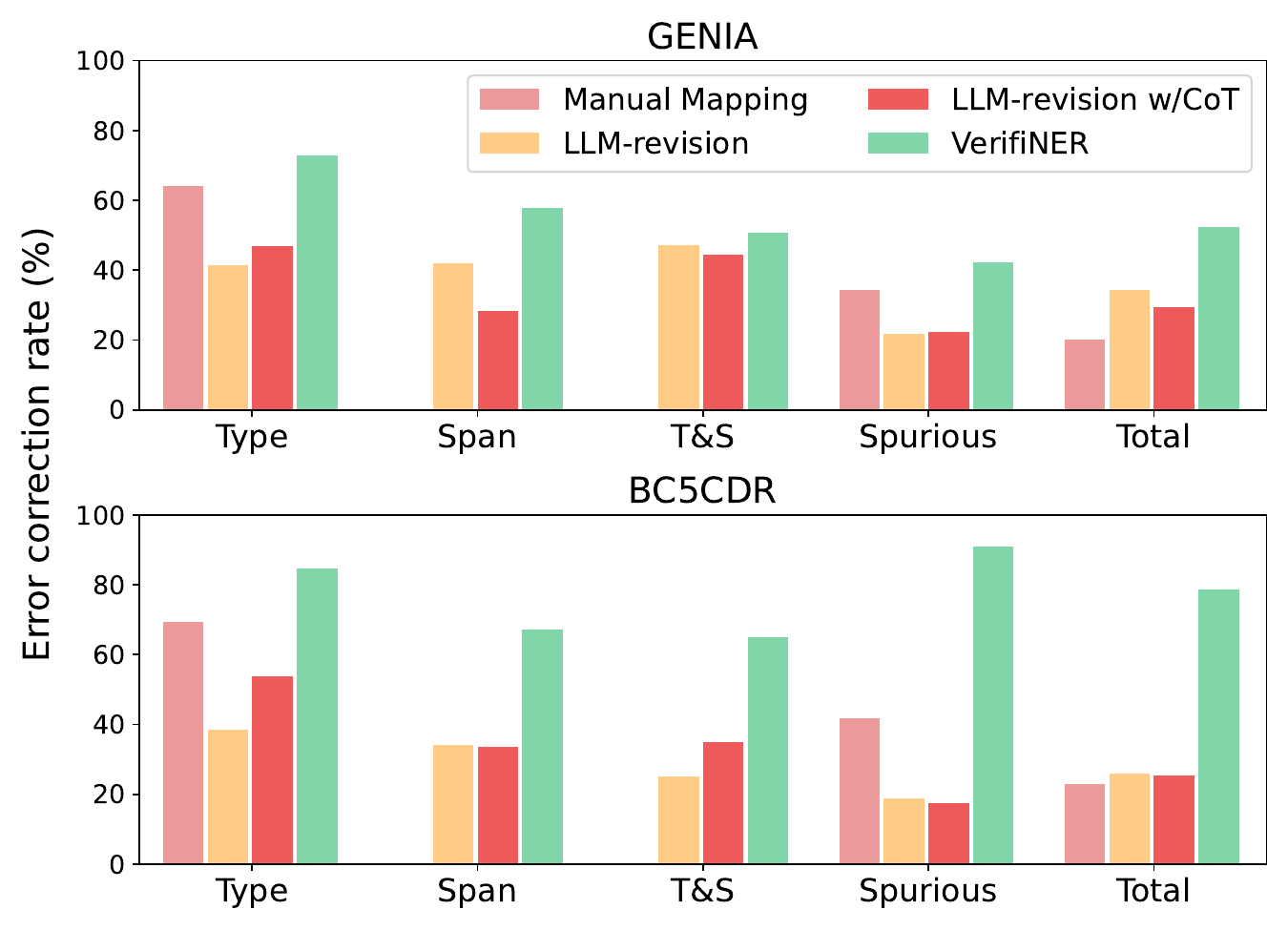}
    
    \caption{Error correction performance of \verifiner compared to other revision baselines. We report averaged rates over all three NER models. Note that the results for Manual Mapping on Span and T\&S does not appear as it can not correct span errors.}
    
    \label{fig:error_drop} 
    
\end{figure}

%% file: tables/unseen_only.tex
\begin{table}[!t]
\small
\centering
\resizebox{1.\columnwidth}{!}{
\begin{tabular}{lcccccc}
\toprule
\multirow{2}{*}{Source $\to$ Target}
 &
 \multicolumn{3}{c}{\textbf{BC5CDR} $\to$ \textbf{GENIA}} &
  \multicolumn{3}{c}{\textbf{GENIA} $\to$ \textbf{BC5CDR}} 
   \\ 
  \cmidrule(lr){2-4} \cmidrule(l){5-7}

\textbf{} & 
  P &
  R &
  F1 &
  P &
  R & 
  F1 
  \\ 
\toprule
GPT-NER & 56.44 & 42.15 & 48.26 & 79.84 & 47.48 & 59.55  \\
\midrule\midrule


ConNER & N/A & N/A & N/A & N/A & N/A & N/A \\
\rowcolor[gray]{0.9}
\textbf{+ \verifiner~} & \textbf{58.15} & \textbf{77.42} & \textbf{66.42} & \textbf{76.74} & \textbf{57.42} & \textbf{65.69} \\ \midrule
BioBERT & N/A & N/A & N/A & N/A & N/A & N/A \\ 
\rowcolor[gray]{0.9}
\textbf{+ \verifiner~} & \textbf{66.49} & \textbf{87.25} & \textbf{75.47} & \textbf{77.64} & \textbf{71.17} & \textbf{74.27} \\

\bottomrule

\end{tabular}
}
\caption{Evaluation on the unseen distribution settings. 
}
\label{tab:unseen_only}
\end{table}

%% file: tables/shift_only.tex
\begin{table}[!t]
\small
\centering
\resizebox{1.\columnwidth}{!}{
\begin{tabular}{lcccccc}
\toprule
\multirow{2}{*}{Source $\to$ Target}
 &
  \multicolumn{3}{c}{\textbf{GENIA$^{\prime}$} $\to$ \textbf{GENIA}} &
  \multicolumn{3}{c}{\textbf{BC5CDR$^{\prime}$} $\to$ \textbf{BC5CDR}}
   \\ 
  \cmidrule(lr){2-4} \cmidrule(l){5-7}

\textbf{} & 
  P &
  R &
  F &
  P &
  R & 
  F 
  \\ 
\toprule

GPT-NER & 56.44 & 42.15 & 48.26 & 79.84 & 47.48 & 59.55 \\
\midrule\midrule

ConNER & 69.97 & 94.10 & 80.26 & 81.46 & 89.47 & 85.28 \\ 
\rowcolor[gray]{0.9}
\textbf{+ \verifiner~} & \textbf{74.16} & 90.03 & \textbf{81.48} & \textbf{94.28} & 85.42 & \textbf{89.63} \\ 
\midrule
BioBERT & 35.36 & 64.50 & 45.68 & 61.75 & 69.95 & 65.59 \\ 
\rowcolor[gray]{0.9}
\textbf{+ \verifiner~} & \textbf{78.91} & \textbf{69.68} & \textbf{74.01} & \textbf{94.16} & \textbf{71.65} & \textbf{81.38} \\

\bottomrule

\end{tabular}
}
\caption{Evaluation on the shifted distribution settings. 
}

\label{tab:unseen_shift}
\end{table}

%% file: figure_latex/low_resource.tex
\begin{figure*}[!t]
    \centering
    \includegraphics[width=1\textwidth]{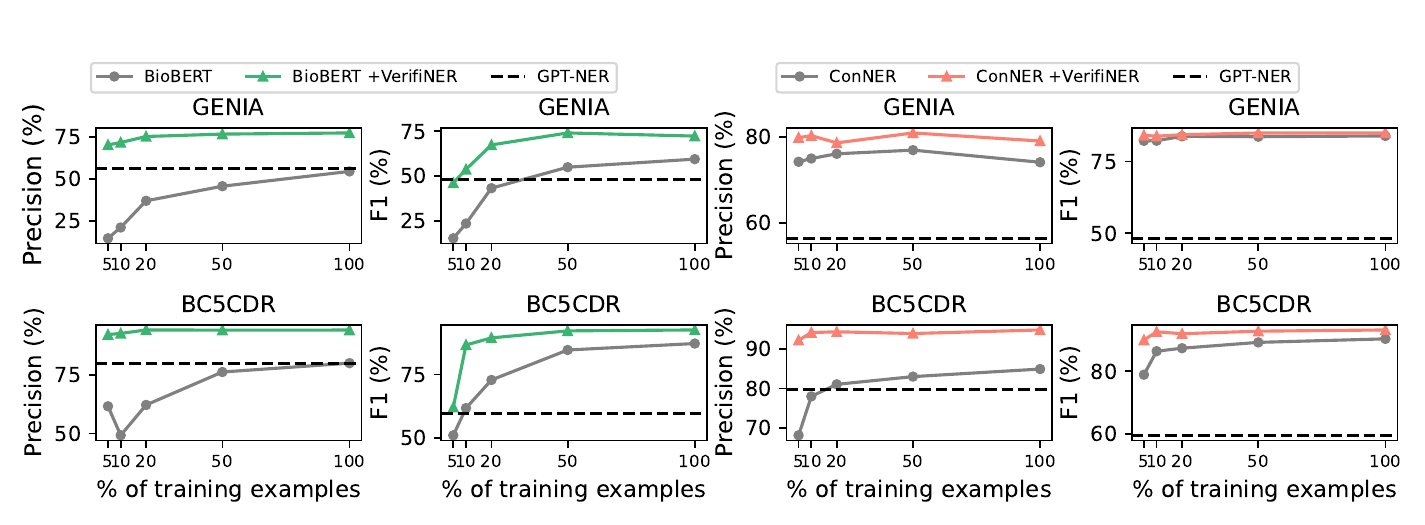}
    \caption{Precision and F1 scores of different NER models (w/ \verifiner) in the low-resource settings. 
    }
    \vspace{-10pt}
    \label{fig:low_resource} ~
\end{figure*} 

%% file: 2_rel.tex
\section{Related Work} 
\paragraph{Named Entity Recognition.}
A conventional approach for solving NER task is to predict probability distribution of entity types and sequentially label each token~\cite{Jeong2023conner, Lee2019biobert}. 
In domain-specific NER, data scarcity challenge has been a consistent challenge. Recent works attempt to solve the issue with data augmentation ~\cite{chen-etal-2021-data-other, zhou-etal-2022-melm-other}.
Other works mitigate the challenge with domain adaptation methods~\cite{zhang-etal-2021-pdaln-other, yang-etal-2022-factmix, xu-etal-2023-improving}.
In most recent years, with the advent of LLMs, a line of work reformulate task definition of NER in a generative manner to leverage LLMs' capabilities in NER tasks. 
\citet{ashok2023promptner} prompts an LLM to produce a list of potential entities with explanations that justify their compatibility with the
provided entity type definitions. 
\citet{wang2023gptner} instructs an LLM to recognize entities by using special tokens to surround them. 

\paragraph{Knowledge-augmented LMs.}
For knowledge augmentation, previous works~\cite{realm, LUKE, retro} show performance improvements on knowledge-intensive tasks. 
\citet{ERNIE} utilize both large-scale corpora and knowledge graphs to train a language model and take full advantage of various knowledge simultaneously.
\citet{ERICA} proposes a novel contrastive learning framework to obtain a deep understanding of the entities and their relations in text.
Similarly, later works~\citep{RAG, kala, control/memory, atals, chae-etal-2023-dialogue} augment LMs with external knowledge during fine-tuning. 
However, they lack in generalization ability across different domains, and require changing model architectures or additional training steps on each task and dataset.

Considering these challenges, recent methods~\cite{internet-augment, cot/augmentation, LLM-Augmenter} use generalization capabilities of large or instruction-tuned LMs to incorporate external knowledge. 
\citet{KAPING} first retrieve relevant facts from a knowledge graph and prepend the retrieved facts to the input question,
which is then forwarded to LLMs to generate the answer.
\citet{zhao-etal-2023-verify} post-edits reasoning chains according to external knowledge to increase prediction factuality. 

\paragraph{NER with External Knowledge.}
There are a number of works that augment external knowledge on NER task. 
\citet{Nie2021} proposes a KB-aware NER framework which incorporates KB knowledge to via type projection to alleviate heterogeneity between KB knowledge and NER type scheme. 
\citet{fu-etal-2023-biomedical} proposes a dictionary-based method that recognizes biomedical entities from input text via synonym generalization.
However, such dictionary-based methods yet fail to generalize over multiple domains.


%% file: 6_conc.tex
\section{Conclusion}
In this paper, we present \verifiner, a verification framework that identifies errors from existing NER methods and revises them into more faithful predictions via knowledge-grounded reasoning. It is the first attempt to solve NER with verification. Specifically, we introduce \verifiner, a plug-and-play verification module which verifies the factuality and contextual relevance of entity with knowledge-grounded reasoning.  
Through extensive experiments, we demonstrate effectiveness of our framework over baselines. It is worth noting that \verifiner is also robust in out-of-domain settings and low-resource scenarios with remarkable performance. 
Considering the feasibility of \verifiner as a model-agnostic approach and its demonstrated generalization capabilities, we expect our work will enhance reliability in NER and be applicable to other domains in future research.

%

%% file: 7_limit.tex
\section*{Limitation}

Despite the remarkable performance of \verifiner, there is still room for improvements on our framework. 
First of all, we conduct experiments using \texttt{gpt-3.5-turbo-1106} model from OpenAI for LLM. Considering the significance of LLM as a major component in the framework, it is yet to be discovered whether the effectiveness of \verifiner is valid with other LLMs. Thus we leave application of our framework on different open and closed LLMs for future work.

In this work, we confine the scope of our work to biomedical domain. 
However, as a plug-and-play framework that can be applied to any model, \verifiner has a potential on other knowledge-intensive domains such as legal or scientific domains. In the future, we plan to investigate its application on other domains. 

Lastly, our framework necessitates multiple inferences using LLMs, which can be computationally expensive. This aspect becomes particularly significant when integrating our framework into applications that require real-time inference. We believe this limitation can be resolved by using smaller models which are distilled from LLMs.

\section*{Ethical Consideration}
The main aspect of our work with the potential for ethical pitfalls is the use of LLMs for our framework. Recent work has highlighted the risks of LLMs in hallucination~\citep{zhang2023siren}. This problem might be more critical in biomedical NER, where prediction can be used for clinical decision support, drug discovery, and personalized medicine. Inaccurate or hallucinated information could lead to misdiagnoses, inappropriate treatment recommendations, or erroneous scientific research directions, ultimately posing a significant risk to patient safety and public health.

We argue here that this risk is largely mitigated in our work, mainly due to our verification process that  incorporates knowledge from KBs. The goal of our work is reducing such errors with verification, specifically by integrating structured data and validated information from KBs. This integration allows for cross-referencing and validation of the outputs generated by the LLM, ensuring that any identified entities and associated information align with reliable biomedical knowledge. 


\section*{Acknowledgements}

This work was supported by the IITP grant funded by the Korea government (MSIT) (No. RS-2020-II201361) and the NRF grant funded by the Korea government (MSIT) (No. RS-2023-00244689).

%% file: 8_appendix.tex
\section{Implementation Details}
\label{sec:appendix}



\label{sec:appendix_implemetation_details}
\paragraph{GENIA Preprocessing.}
Although GENIA is a nested NER benchmark, we choose GENIA to investigate the performance of the flat NER models because 1) we wanted to validate the robustness of our framework on datasets with more labels than BC5CDR, and 2) there are a limited number of available biomedical NER datasets. To fine-tune GENIA on models that are trained for flat NER, we transform GENIA into a flat NER task by separating a single instance with nested labels into multiple individually labeled instances.

\paragraph{Dataset Statistics and Sampling.}
 Due to the budget constraints, we use randomly sampled test set from both GENIA and BC5CDR for all experiments. 
 We randomly sample 500 and 100 documents from GENIA and BC5CDR, respectively. Then, to verify each entity, we reconstruct context documents into entity-level input context. Specifically, we divide document per entity and verify each. 
 We provide the detailed dataset statistics in Table ~\ref{tab:data_stat}, and detailed error ratio of the models in ~\ref{tab:pie_stat}.
\input{tables/dataset_statistics}

\input{tables/pie_stat}
\paragraph{Baselines.}
To apply \verifiner, we consider two groups of NER models as baselines: the fine-tune based NER model (\ie{} ConNER and BioBERT), and prompting-based LLM (\ie{} GPT-NER). 
In case of ConNER and GPT-NER, we follow the official implementation of ~\citep{Jeong2023conner} and ~\citep{wang2023gptner}, respectively. 
In case of BioBERT ~\citep{Lee2019biobert}, we use the checkpoint of pre-trained model from huggingface\footnote{https://huggingface.co/dmis-lab/biobert-v1.1}. 
We train BioBERT on both datasets (~\ie{} GENIA and BC5CDR) for 20 epochs with learning rate of 3e-5. 

\paragraph{KB Augmentation.}
Inspired from the training process of language models that learn reliable knowledge from the train set, we add entities from the train set to KB as reliable knowledge. It is a fair setting because the entities are not directly assigned as the answer because not only the exact span, but also other multiple spans are given as candidates, and LLM has to choose the answer among them. 

In order to set a scenario where all gold entities exist completely in the knowledge source, we augment the knowledge base (KB) using entities present in the dataset. First, we search all possible candidate spans extracted from initial prediction to the UMLS database. 
We then add these results to the dictionary if the results were found from the database. 
Subsequently, for entities that could not be found in the search, we additionally search within the annotated entity set present in each dataset (GENIA, BC5CDR) and add them to the dictionary.
Through this process, we create a complete external knowledge source containing over 90\% of the knowledge from the gold entities.

It is worth noting that our framework is applicable even when not all knowledge is in the KB or when new entities emerge, because as long as there exists a KB that contains knowledge of the query entity (no matter if the KB is complete or not) and we can retrieve knowledge, VerifiNER is able to utilize knowledge and post-edit NER errors.

\paragraph{Manual Mapping.}
We provide the statistics of the number of semantic types assigned to each label.
\input{tables/manual_mapping}


\input{tables/shift_2}

\input{figure_latex/subset_distribution}

\section{Experimental Settings}
\label{ssec:appendix_exp_settings}

\paragraph{Unseen Distribution Settings.}
We fine-tune \textsc{ConNER} and \textsc{BioBERT} using the training sets of the GENIA and BC5CDR datasets, respectively.
Then using the checkpoints from each trained model, we cross-infer on the test set of remaining dataset, which is not used during train phase.

When applying \verifiner to the inferred predictions, we adjust the initial predictions by  a threshold to the logits of the model output. 
This step is implemented to mitigate the occurrence of an excessive number of "O" tags in the generated output, which could otherwise lead to a scarcity of entity input to the verification.
Specifically, we compute the output logit percentile for the "O" tags across all instances. 
Then, for the top 10\% of "O" tokens, we retain the original logits, while for the rest, we multiply the logits by -9999. 
This adjustment ensure that only predictions confidently classified as "O" tags remained, while the rest are assigned as entities for the input of verification.
\paragraph{Shifted Distribution Settings.}
To assume the scenarios where the target distributions differ from the source distributions, we manually split both datasets into two distinct subsets. 
We then fine-tune \textsc{ConNER} and \textsc{BioBERT} using each subset as a source and inferred on each test set. 
The distribution of each subset and the experimental results for subset 2 are presented in Figure~\ref{fig:label_dist_GENIA} and Table~\ref{tab:shift_2}, respectively.

\paragraph{Computational Resources and API Cost.}
We run ConNER and BioBERT on eight NVIDIA RTX A5000 GPUs. For ChatGPT API usage, we use \$ 470 in total. 




\section{Case Study}
\label{sec:case_study}
We select representative examples of verification result by \verifiner from span error case in GENIA and t\&s error case in BC5CDR.

Table~\ref{tab:case_genia} presents a span error verification from GENIA. The two candidates are very similar with each other, yet \verifiner manages to distinguish the answer between the two. Specifically, \verifiner selects the gold annotation ``\textit{human lymphocytes}'' as the final answer, based on the evidence that ``\textit{human lymphocytes}'' narrows down the scope to human cells specifically. 

Table~\ref{tab:case_bc5} shows a type \& span error verification result from BC5CDR. It is evident that the intitial prediction, ``\textit{bupivacaine arrhythmogenicity}'' is not in the candidate list. This indicates that the NER model has predicted an invalid entity that is not registered in the KB. Luckily, \verifiner is able to filter out such invalid entities by searching the candidates in KB and verify their correctness with using knowledge as evidence.

\section{Prompts}
\label{sec:prompts}

We provide type factuality verification prompts in Table~\ref{tab:type_prompt_genia},~\ref{tab:type_prompt_bc5}, and contextual relevance verification prompts in Table~\ref{tab:con_prompt_genia},~\ref{tab:con_prompt_bc5}.
Note that we use individual prompt for GENIA and BC5CDR respectively. 

\input{tables/case_study}
\input{tables/step2_prompt}

\input{tables/step3_prompt}

\input{tables/prompt_baseline}

%% file: tables/dataset_statistics.tex



\newcolumntype{L}{>{\raggedright\arraybackslash}p{0.2\columnwidth}}
\newcolumntype{C}{>{\centering}p{0.1\columnwidth}}


\begin{table}[!ht]
\centering
\resizebox{1.\columnwidth}{!}{
\begin{tabular}{lccccc}
\toprule
Dataset & Domain & Types & Split & \#Document & \#Entities \\ \midrule
\multirow{2}{*}{GENIA} &
  \multirow{2}{*}{biomedical} &
  \multirow{2}{*}{\makecell{protein, RNA, DNA,\\ cell\_type, cell\_line}} &
  train &
  16615 &
  44488 \\
        &        &   &         test     & 500        & 1472       \\
    \midrule
\multirow{2}{*}{BC5CDR} &
  \multirow{2}{*}{biomedical} &
  
  \multirow{2}{*}{\makecell{Chemical, Disease}} &
  train &
  500 &
  9366 \\
        &        &   &       test      & 100        & 1831       \\ 

\bottomrule

\end{tabular}
}
\caption{Dataset Statistics}
\label{tab:data_stat}
\end{table}

%% file: tables/pie_stat.tex
\begin{table}[!ht]
\small
\centering
\resizebox{1.\columnwidth}{!}{
\begin{tabular}{lcccccc}
\toprule
Model   &   Type    &   Span    &   Type\&Span  &   Spurious & FP & FN \\
\midrule
ConNER  &   10.85   &   32.77   &   8.51        &   38.94   &   91.1    &   8.9 \\
BioBERT &   2.39    &   33.61   &   2.9         &   22.2    &   61.1    &   38.9 \\
GPT-NER &   4.22    &   11.71   &   4.92        &   15.14   &   64  &   36 \\
\bottomrule
\end{tabular}
}
\caption{Ratio of error types of NER models.}
\label{tab:pie_stat}
\end{table}

%% file: tables/manual_mapping.tex
\begin{table}[!h]
\small
\centering
{
\begin{tabular}{cc|c}
\toprule
Chemical   &   Disease    &   Total \\
\midrule
13  &   15   &   28 \\
\bottomrule
\end{tabular}
}
\caption{Distribution of semantic types mapped into BC5CDR labels.}
\label{tab:bc5_map}
\end{table}

\begin{table}[!h]
\small
\centering
\resizebox{1.\columnwidth}{!}{
\begin{tabular}{ccccc|c}
\toprule
Protein & Cell Type	& DNA & Cell Line & RNA & Total \\
\midrule
27 & 10 & 18 & 8 & 5 & 68 \\
\bottomrule
\end{tabular}
}
\caption{Distribution of semantic types mapped into GENIA labels.}
\label{tab:genia_map}
\end{table}

%% file: tables/shift_2.tex
\begin{table}[!t]
\small
\centering
\resizebox{1.\columnwidth}{!}{
\begin{tabular}{lcccccc}
\toprule
\multirow{2}{*}{Source $\to$ Target}
 &
  \multicolumn{3}{c}{\textbf{GENIA$^{\prime}$} $\to$ \textbf{GENIA}} &
  \multicolumn{3}{c}{\textbf{BC5CDR$^{\prime}$} $\to$ \textbf{BC5CDR}}
   \\ 
  \cmidrule(lr){2-4} \cmidrule(l){5-7}

\textbf{} & 
  P &
  R &
  F &
  P &
  R & 
  F 
  \\ 
\toprule
\textsc{GPT-NER} & 56.44 & 42.15 & 48.26 & 79.84 & 47.48 & 59.55 \\
\midrule\midrule




\textsc{ConNER
} & 74.38 & 59.57 & 66.16 & 75.73 & 84.85 & 80.03 \\ 
\textbf{+ \verifiner~} & \textbf{78.86} & \textbf{59.8} & \textbf{68.02} & \textbf{74.16} & 90.03 & \textbf{81.48} \\ 
\midrule
\textsc{BioBERT
} & 15.16 & 16.12 & 15.62 & 46.45 & 55.49 & 50.57 \\ 
\textbf{+ \verifiner~} & \textbf{78.27} & \textbf{38.77} & \textbf{51.85} & \textbf{93.33} & \textbf{61.92} & \textbf{74.45} \\

\bottomrule

\end{tabular}
}
\caption{Experimental results on distribution shift settings. 
}
\label{tab:shift_2}
\end{table}

%% file: figure_latex/subset_distribution.tex
\begin{figure}[!t]
    \centering
    \includegraphics[width=1\columnwidth]{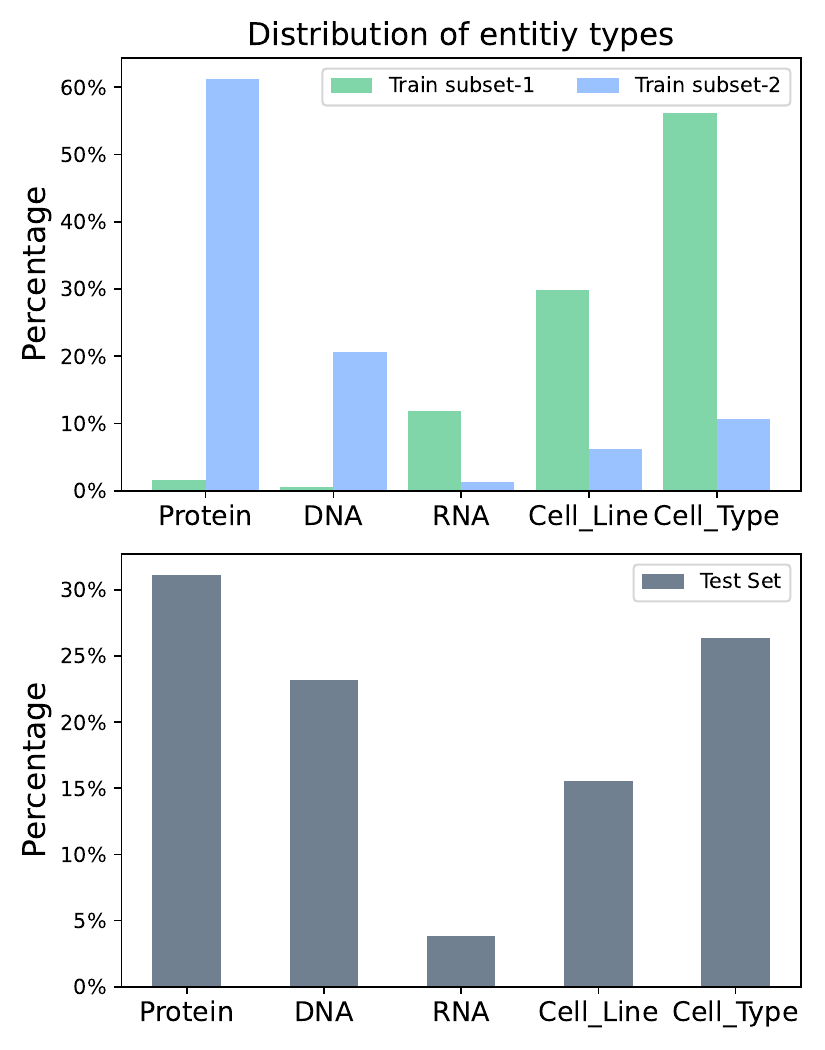}
    
    \caption{Label distribution of splitted GENIA train subsets and test set for shifted distribution experiments.}
    
    \label{fig:label_dist_GENIA} 
    
\end{figure} 

%% file: tables/case_study.tex
\begin{table*}
    \small
    \centering
    \resizebox{1\textwidth}{!}{
        \begin{tabular}{p{0.18\textwidth}p{0.8\textwidth}>{\centering\arraybackslash}p{0.02\textwidth}}
            \toprule
            \multicolumn{2}{l}{\textbf{Example of Span Error Correction from GENIA}} \\
            \midrule
            Initial prediction  & (`lymphocytes', `cell\_type') & \\
            \midrule
            Context             & At 10 microM, both compounds inhibited IL-2 mRNA and protein levels in the NFAT-1-linked lac-Z transfectants, and in human lymphocytes. \\
            \midrule
            Candidate pairs      & {
                \textbf{[Pair 1]}

                Candidate: `lymphocytes'

                Knowledge: 

                Evidence: `Based on the given knowledge, the semantic type of lymphocytes is `cell\_type'. Lymphocytes are a type of white blood cell that plays a crucial role in the immune system. They are responsible for recognizing and attacking foreign substances in the body. In the given paragraph, lymphocytes are mentioned in the context of inhibiting IL-2 mRNA and protein levels. IL-2 is an important cytokine involved in immune responses, and its inhibition suggests a potential role of lymphocytes in regulating immune activity. Therefore, based on the semantic type and the context, the class of the entity lymphocytes is cell\_type.'

                \textbf{[Pair 2]}

                Candidate: `human lymphocytes'

                Knowledge: 

                Evidence: `Based on the given knowledge, the semantic type of human lymphocytes is `cell\_type'. Lymphocytes are a type of white blood cell that plays a crucial role in the immune system. In the given paragraph, human lymphocytes are mentioned in the context of inhibiting IL-2 mRNA and protein levels. This aligns with the role of lymphocytes in immune responses and their involvement in cytokine production. Therefore, the class of the entity `human lymphocytes' is cell\_type.'
            } & \\
            \midrule
            Reasoning paths      & {
                \textbf{[Reasoning Path 1]}

                Based on the provided explanations, both candidate entities `lymphocytes' and `human lymphocytes' are described as types of white blood cells that play a crucial role in the immune system. They are both mentioned in the context of inhibiting IL-2 mRNA and protein levels, which aligns with their involvement in immune responses and cytokine production. 

                However, the specific mention of "human lymphocytes" in the paragraph suggests a focus on human cells specifically. While lymphocytes in general can include various types of cells, \textcolor{magenta}{the mention of "human lymphocytes" narrows down the scope to human cells specifically.} This specificity makes "human lymphocytes" the more appropriate and specific entity in the given biomedical context. 

                Therefore, the final answer is (`human lymphocytes', cell\_type).

                \textbf{[Reasoning Path 2]}

                Based on the provided explanations, both candidate entities `lymphocytes' and `human lymphocytes' are described as a type of white blood cell that plays a crucial role in the immune system. They are both mentioned in the context of inhibiting IL-2 mRNA and protein levels. Although the term `human lymphocytes' provides more specificity, it is important to note that the given knowledge does not provide any additional information distinguishing between lymphocytes and human lymphocytes. Therefore, it is not possible to definitively determine which term is the most reliable entity in this context. 
                Therefore, the final answer is (None, None).

                \textbf{[Reasoning Path 3]}

                ...
            } & \\
            \midrule
            Votes & {
                (`human lymphocytes', `cell\_type') 8

                (`lymphocytes', `cell\_type') 1

                (None, None) 1
            } & \\
            \midrule
            Corrected prediction & (`human lymphocytes', `cell\_type') & \\
            \midrule
            Gold annotation & (`human lymphocytes', `cell\_type') & \\
            \bottomrule
        \end{tabular}
    }
    \caption{Example from GENIA, span error is corrected.}
    \label{tab:case_genia}
\end{table*}

\begin{table*}
    \small
    \centering
    \resizebox{1\textwidth}{!}{
        \begin{tabular}{p{0.18\textwidth}p{0.8\textwidth}>{\centering\arraybackslash}p{0.02\textwidth}}
            \toprule
            \multicolumn{2}{l}{\textbf{Example of Type\&Span Error Correction from BC5CDR}} \\
            \midrule
            Initial prediction  & (`bupivacaine arrhythmogenicity', `Disease') & \\
            \midrule
            Context             & Progesterone potentiation of bupivacaine arrhythmogenicity in pentobarbital-anesthetized rats and beating rat heart cell cultures. \\
            \midrule
            Candidate pairs      & {
                (`arrhythmogenicity', `Arrhythmogenicity is a congenital cardiomyopathy characterized by the infiltration of adipose and fibrous tissue into the right ventricle wall and the loss of myocardial cells. It primarily affects the free wall of the right ventricle and right atria, resulting in ventricular and supraventricular arrhythmias. Based on this definition and the semantic type of `arrhythmogenicity' is `Disease or Syndrome'.)

                (`bupivacaine', `Bupivacaine is a widely used local anesthetic agent. It falls under the semantic type of `Organic Chemical' and `Pharmacologic Substance'. Based on this information, `bupivacaine' can be classified as `Chemical'.)
            } & \\
            \midrule
            Reasoning paths      & {
                \textbf{[Reasoning Path 1]}

                Based on the given knowledge, 'bupivacaine' is a widely used local anesthetic agent and is classified as a Chemical entity. On the other hand, 'arrhythmogenicity' is a congenital cardiomyopathy characterized by the infiltration of adipose and fibrous tissue into the right ventricle wall and the loss of myocardial cells. It primarily affects the free wall of the right ventricle and right atria, resulting in ventricular and supraventricular arrhythmias. Based on this definition and the semantic type of 'Disease or Syndrome', arrhythmogenicity can be classified as a disease. \textcolor{magenta}{Comparing these entities, 'bupivacaine' is directly related to the context as a local anesthetic agent,} and it satisfies the guideline of annotating the most specific chemical mentions. 

                Therefore, the most appropriate answer entity is 'bupivacaine'.

                \textbf{[Reasoning Path 2]}

                Based on the given knowledge, both 'arrhythmogenicity' and 'bupivacaine' are classifiable entities. 'Arrhythmogenicity' is a congenital cardiomyopathy characterized by the infiltration of adipose and fibrous tissue into the right ventricle wall and the loss of myocardial cells. It primarily affects the free wall of the right ventricle and right atria, resulting in ventricular and supraventricular arrhythmias. Based on this definition and the semantic type of 'Disease or Syndrome', arrhythmogenicity can be classified as a disease. On the other hand, 'bupivacaine' is a widely used local anesthetic agent and falls under the semantic type of 'Organic Chemical' and 'Pharmacologic Substance', indicating that it is a chemical. Comparing the two entities, they represent distinct classes of entities based on their definitions and semantic types. 

                Therefore, the final answer is ('arrhythmogenicity', Disease).

                \textbf{[Reasoning Path 3]}

                ...
            } & \\
            \midrule
            Votes & {
                (`bupivacaine', `Chemical') 9

                (`arrhythmogenicity', `Disease') 1
            } & \\
            \midrule
            Corrected prediction & (`bupivacaine', `Chemical') & \\
            \midrule
            Gold annotation & (`bupivacaine', `Chemical') & \\
            \bottomrule
        \end{tabular}
    }
    \caption{Example from BC5CDR, type\&span error is corrected.}
    \label{tab:case_bc5}
\end{table*}

%% file: tables/step2_prompt.tex
\begin{table*}
    \small
    \centering
    \begin{tabular}{p{14cm}}
    \toprule
    \textbf{Type Factuality Verification Prompt (GENIA)} \\
    \midrule
\textcolor{teal}{\textbf{[Task Description]}}\\
You are solving named entity recognition task for biomedical articles. There is an entity to which you need to assign either one of the six classes : protein, DNA, RNA, cell\_line, cell\_type or None. Class None is assigned when an entity does not belong to any of the five classes: protein, DNA, RNA, cell\_line, and cell\_type. I will give you an entity and reliable knowledge about the entity for reference. There are two types of knowledge: definition and semantic type of entities you need to predict. Identify the class of the entity and explain why. When generating explanation, you need to quote the knowledge. If both definition and semantic type are not provided, use context information of entity in Paragraph or your own knowledge to explain.
Remember, you should assign at least one class for each entity from above six (protein, DNA, RNA, cell\_line, cell\_type or None). Using Example 1 to 6 as reference, answer Example 7.
    \\\\

    \textcolor{teal}{\textbf{[Example 1]}} \\
\textbf{Entity:} GRalpha
\\
\textbf{Knowledge about this entity:}
The definition of GRalpha is not provided.
The semantic type of GRalpha is [`Amino Acid, Peptide, or Protein', `Receptor'].
\\
\textbf{Explanation:} Based on the given knowledge, the semantic type of GRalpha is `Amino Acid, Peptide, or Protein' and `Receptor'. Proteins are large molecules that perform various functions in the body, including acting as receptors for hormones and other signaling molecules. In this case, GRalpha is described as a hormone-activated transcription factor, which aligns with its role as a protein receptor. Therefore, the class of the entity GRalpha is protein. \\\\

    \textbf{\textcolor{teal}{\textbf{[Example 2]}}} \\
\textbf{Entity:} neutrophils
\\
\textbf{Knowledge about this entity:}\\
The definition of neutrophils is [`Granular leukocytes having a nucleus with three to five lobes connected by slender threads of chromatin, and cytoplasm containing fine inconspicuous granules and stainable by neutral dyes.'].
The semantic type of neutrophils is [`Cell'].
\\
\textbf{Explanation:}\\
Based on the given knowledge, the definition of neutrophils is `Granular leukocytes having a nucleus with three to five lobes connected by slender threads of chromatin, and cytoplasm containing fine inconspicuous granules and stainable by neutral dyes.' This definition indicates that neutrophils are a type of cell, specifically a type of granular leukocyte. Therefore, the class of the entity neutrophils is cell\_type.\\\\

    \textcolor{teal}{\textbf{[Example 3]}} ... \\ \bottomrule
    \end{tabular}
    \caption{The prompt for type factuality verification on GENIA (Examples 3 to 6 are omitted in this table).}
    \label{tab:type_prompt_genia}
\end{table*}


\begin{table*}
    \small
    \centering
    \begin{tabular}{p{14cm}}
    \toprule
    \textbf{Type Factuality Verification Prompt (BC5CDR)} \\
    \midrule
\textcolor{teal}{\textbf{[Task Description]}}\\
You are solving named entity recognition task for biomedical articles. There is an entity to which you need to assign either one of the three classes: Chemical or Disease or None. You should only use one of the three classes. No other class exists. I will give you an entity and reliable knowledge about the entity. There are two types of knowledge: definition and semantic type. Identify the class of the entity and explain why. When generating explanation, you need to use the given knowledge. If you are not certain with your decision, you may explain why you are not certain. Even if it does not fit to any of the classes, try to assign the most related one. 
    \\\\

    \textcolor{teal}{\textbf{[Example 1]}} \\
    \textbf{Paragraph:}\\
\textbf{Entity:} chest pain
\\
\textbf{Knowledge about this entity:}
The definition of chest pain is ['Pressure, burning, or numbness in the chest.'].
The semantic type of chest pain is ['Sign or Symptom'].
\\
\textbf{Explanation:} 
Chest pain is a symptom characterized by pressure, burning, or numbness in the chest. Based on the provided knowledge, chest pain falls under the semantic type of 'Sign or Symptom'. Therefore, the class of the entity chest pain is Disease.
\\\\

    \textbf{\textcolor{teal}{\textbf{[Example 2]}}} \\
\textbf{Entity:} amoxicillin
\\
\textbf{Knowledge about this entity:}\\
The definition of amoxicillin is ['A broad-spectrum semisynthetic antibiotic similar to AMPICILLIN except that its resistance to gastric acid permits higher serum levels with oral administration.'].
The semantic type of amoxicillin is ['Organic Chemical', 'Antibiotic'].
\\
\textbf{Explanation:}\\
amoxicillin is a broad-spectrum semisynthetic antibiotic that is similar to AMPICILLIN. It has the ability to resist gastric acid, allowing for higher serum levels with oral administration. Additionally, it is classified as an organic chemical and an antibiotic. Based on this information, it can be concluded that amoxicillin belongs to the class of Chemical. Therefore, the class of the entity amoxicillin is Chemical.
\\\\

    \textcolor{teal}{\textbf{[Example 3]}} ... \\ \bottomrule
    \end{tabular}
    \caption{The prompt for type factuality verification on BC5CDR (Examples 3 to 6 are omitted in this table).}
    \label{tab:type_prompt_bc5}
\end{table*}

%% file: tables/step3_prompt.tex
\begin{table*}
    \small
    \centering
    \begin{tabular}{p{14cm}}
    \toprule
    \textbf{Contextual Relevance Verification Prompt (GENIA)} \\
    \midrule
\textbf{[Task Description]}\\
You are solving named entity recognition task for biomedical articles. You will be provided pairs of candidate entities and explanations of why each entity is labeled such, along with a context in which the candidate entities appear. The entities are very similar, and your job is to choose a pair that has the most reliable explanation, and is most likely to be a biomedical entity considering the context. Think Step-by-step and explain why you think that entity is the answer and compare your answer entity to other candidate entities. Using Example 1 to 8 as reference, answer Example 9.
    \\\\
\textcolor{teal}{\textbf{[Example 1]}} \\
\textbf{Pair of candidate entity and explanation:}\\
{[`proto-oncogene c-rel', `Since the definition and semantic type of proto-oncogene c-rel are not provided, we can use our own knowledge to determine the class of the entity. Proto-oncogenes are genes that have the potential to become oncogenes, which are genes that can cause cancer when mutated or overexpressed. In this case, proto-oncogene c-rel is mentioned in the context of a DNA probe that spans a conserved domain among other genes, including the p50 DNA binding subunit of NF-kappa B. NF-kappa B is a transcription factor that plays a key role in regulating genes involved in immune responses and cell survival. The fact that proto-oncogene c-rel is mentioned in relation to DNA binding suggests that it may also be involved in gene regulation. Therefore, based on our knowledge, the class of the entity proto-oncogene c-rel is protein.']}\\
{[`c-rel', `In the given paragraph, c-rel is mentioned as a proto-oncogene that is conserved among other genes, including the p50 DNA binding subunit of NF-kappa B. Proto-oncogenes are genes that have the potential to cause cancer when mutated or overexpressed. In this case, c-rel is described as a proto-oncogene, indicating its role in cancer development. Therefore, the class of the entity c-rel is protein.']}
\\
\textbf{Your decision:}\\
Based on the provided explanation, `proto-oncogene c-rel' links it to a conserved domain among other genes, including the p50 DNA binding subunit of NF-kappa B. The explanation goes on to connect proto-oncogene c-rel to the role of proto-oncogenes in cancer development, suggesting its involvement in gene regulation. This detailed explanation provides a broader understanding of the entity in the given biomedical context. In comparison, the other candidate entity `c-rel' is less specific in its explanation, merely stating that c-rel is a proto-oncogene without providing the additional context about its conserved domain and its association with the p50 DNA binding subunit of NF-kappa B. The chosen entity, `proto-oncogene c-rel,' is more comprehensive and aligns better with the information provided in the paragraph. \\
Therefore, the final answer is (`proto-oncogene c-rel' , protein).
\\\\

    \textbf{\textcolor{teal}{\textbf{[Example 2]}}} \\
\textbf{Pair of candidate entity and explanation:}\\
{[`CBF', `Based on the given knowledge, the semantic type of CBF is protein. Proteins are large molecules that perform various functions in the body. In the given paragraph, CBF is described as consisting of two subunits, CBF alpha and CBF beta, with CBF alpha being a DNA binding subunit. This aligns with the role of proteins as molecules that can bind to DNA. Therefore, based on the semantic type information and the context provided, the class of the entity CBF is protein.']}\\
{[`CBF beta', `Based on the given knowledge, the semantic type of CBF beta is protein. In the given paragraph, CBF beta is described as a subunit that stimulates the DNA binding activity of CBF alpha. Therefore, the class of the entity CBF beta is protein.']}\\
{[`beta', 'Based on the given knowledge, the semantic type of beta is 'protein'. Proteins are large molecules that perform various functions in the body. In this case, 'beta' is mentioned as a subunit that stimulates the DNA binding activity of CBF alpha, indicating its role as a protein. Therefore, the class of the entity `beta' is protein.']}
\\
\textbf{Your decision:}\\
Based on the provided explanations, candidate `CBF beta' is the subunit of CBF, but `CBF beta' is specifically highlighted as a subunit that stimulates the DNA binding activity of CBF. The explanation for `CBF beta' explicitly assigns it the semantic type of protein, aligning with the general role of proteins in the body. It is a specific subunit with a clear role in stimulating DNA binding activity, and its semantic type aligns with the expected classification of proteins.\\
Therefore, the final answer is (`CBF beta', protein).
    \\\\
    
    \textcolor{teal}{\textbf{[Example 3]}} 
    ... \\ \bottomrule
    \end{tabular}
    \caption{The prompt for contextual relevance verification on GENIA (Examples 3 to 8 are omitted in this table).}
    \label{tab:con_prompt_genia}
\end{table*}

\begin{table*}
    \small
    \centering
    \begin{tabular}{p{14cm}}
    \toprule
    \textbf{Contextual Relevance Verification Prompt (BC5CDR)} \\
    \midrule
\textbf{[Task Description]}\\
You are solving named entity recognition task for biomedical articles. You will be provided pairs of candidate entities and explanations of why each entity is labeled such, along with a context in which the candidate entities appear. The entities are very similar, and your job is to choose a pair that has the most reliable explanation, and is most likely to be a biomedical entity considering the context. Explain why you think that entity is the answer and compare your answer entity to other candidate entities. Use Example 1 to 20 as reference.
    \\\\
    
\textcolor{teal}{\textbf{[Example 1]}} \\
\textbf{Pair of candidate entity and explanation:}\\
{[`hepatitis', `hepatitis is defined as inflammation of the liver. Based on this definition, it can be classified as a disease. Additionally, the semantic type of hepatitis is also `Disease or Syndrome'. Therefore, the entity hepatitis should be classified as a Disease.']}\\
{[`fulminant hepatitis', `Since there is no definition available for fulminant hepatitis, I will rely solely on the semantic type. The semantic type of fulminant hepatitis is 'Disease or Syndrome', indicating that it is a disease. 
Therefore, I would classify fulminant hepatitis as a Disease.']}\\

\textbf{Your decision:}\\
Based on the given knowledge, hepatitis is defined as inflammation of the liver, which falls under the category of a disease. Additionally, the semantic type of hepatitis is also `Disease or Syndrome'. Therefore, the entity hepatitis should be classified as a Disease. On the other hand, there is no specific definition available for fulminant hepatitis. However, the semantic type of fulminant hepatitis is `Disease or Syndrome', indicating that it is a disease. Therefore, both entities can be classified as Disease. However, `fulminant hepatitis' provides additional information about the severity of the hepatitis, indicating a rapid and severe progression. This makes it a more specific mention compared to `hepatitis'. \\
Therefore, the final answer is (`fulminant hepatitis', Disease).
\\\\

    \textbf{\textcolor{teal}{\textbf{[Example 2]}}} \\
\textbf{Pair of candidate entity and explanation:}\\
{[`haloperidol', `The definition of haloperidol is `A phenyl-piperidinyl-butyrophenone that is used primarily to treat SCHIZOPHRENIA and other PSYCHOSES. It is also used in schizoaffective disorder, DELUSIONAL DISORDERS, ballism, and TOURETTE SYNDROME (a drug of choice) and occasionally as adjunctive therapy in INTELLECTUAL DISABILITY and the chorea of HUNTINGTON DISEASE. It is a potent antiemetic and is used in the treatment of intractable HICCUPS. (From AMA Drug Evaluations Annual, 1994, p279) 
The semantic type of haloperidol is `Organic Chemical', `Pharmacologic Substance'.]}\\
\textbf{Your decision:}\\
haloperidol is an organic chemical and pharmacologic substance that is primarily used to treat schizophrenia and other psychoses. It is also used in schizoaffective disorder, delusional disorders, ballism, Tourette syndrome, and occasionally as adjunctive therapy in intellectual disability and the chorea of Huntington disease. It is a potent antiemetic and is used in the treatment of intractable hiccups. Based on the provided knowledge, haloperidol is classified as a pharmacologic substance. \\
Therefore, the final answer is (`haloperidol', Chemical). \\
    \\\\
    
    \textcolor{teal}{\textbf{[Example 3]}} 
    ... \\ \bottomrule
    \end{tabular}
    \caption{The prompt for contextual relevance verification on BC5CDR (Examples 3 to 20 are omitted in this table).}
    \label{tab:con_prompt_bc5}
\end{table*}

%% file: tables/prompt_baseline.tex
\begin{table*}
    \small
    \centering
    \begin{tabular}{p{14cm}}
    \toprule
    \textbf{LLM-revision Prompt (GENIA)} \\
    \midrule
\textbf{[Task Description]}\\
Within the GENIA dataset consisting of PubMed articles, the sentences are tagged with labels.
The task is to verify whether the word is properly tagged with the correct label from the given sentence and revise it.
If the entity is annotated incorrectly, you may revise tags.
Here’s what each label means: <P> is `protein', <D> is `DNA', <R> is `RNA', <CL> is `cell\_line', <CT> is `cell\_type'.
If the entity corresponds to Protein, <P> should be added before the entity and </P> after it.
Follow the format of Example 1 to answer Example 2. 
\\\\
\textcolor{teal}{\textbf{[Example 1]}} \\
\textbf{The given sentence:}  \\
Whereas activation of the <D>HIV-1 enhancer</D> following T-cell stimulation is mediated largely through binding of the transcription factor <P>NF-kappa B</P> to two adjacent <D>kappa B sites</D> in the <R>HIV-1</R> long terminal repeat , activation of the <D>HIV-2 enhancer</D> in <CT>monocytes</CT> and <CT>T cells</CT> is dependent on four <D>cis-acting elements</D> : a single kappa B site , two purine-rich binding sites , <D>PuB1</D> and <D>PuB2</D> , and a pets site .

\textbf{The labeled sentence:} \\
Whereas activation of the <D>HIV-1 enhancer</D> following T-cell stimulation is mediated largely through binding of the <P>transcription factor</P> <P>NF-kappa B</P> to two adjacent <D>kappa B sites</D> in the <D>HIV-1 long terminal repeat</D> , activation of the <D>HIV-2 enhancer</D> in <CT>monocytes</> and </CT>T cells</> is dependent on four <D>cis-acting elements</D> : a single <D>kappa B site</D> , two purine-rich binding sites , <D>PuB1</D> and <D>PuB2</D> , and a <D>pets site</D> .
\\\\
\textcolor{teal}{\textbf{[Example 2]}} \\
\textbf{The given sentence:} \\
\{context\}
\\
\textbf{The labeled sentence: }
    \end{tabular}
    \caption{The prompt for LLM-revision on GENIA.}
    \label{tab:baseline_prompt_genia}
\end{table*}

\begin{table*}
    \small
    \centering
    \begin{tabular}{p{14cm}}
    \toprule
    \textbf{LLM-revision Prompt (BC5CDR)} \\
    \midrule
\textbf{[Task Description]}\\
Within the BC5CDR dataset consisting of PubMed articles, the sentences are tagged with labels.
The task is to verify whether the word is properly tagged with the correct label from the given sentence and revise it. If the entity is annotated incorrectly, you may revise tags.
Here’s what each label means: <C> is for `Chemical', <D> is for `Disease'.
If the entity corresponds to Disease, <D> should be added before the entity and </D> after it.
Follow the format of Example 1 to answer Example 2.
\\\\
\textcolor{teal}{\textbf{[Example 1]}} \\
\textbf{The given sentence:}  \\
Recurarization in the recovery room . A case of recurarization in the recovery room is reported . Accumulation of <C>atracurium</C> in the intravenous line led to recurarization after flushing the line in the recovery room . A <D>respiratory arrest</D> with severe desaturation and bradycardia occurred . Circumstances leading to this event and the mechanisms enabling a neuromuscular blockade to occur , following the administration of a small dose of relaxant , are discussed .\\
\textbf{The labeled sentence:}\\
Recurarization in the recovery room . A case of recurarization in the recovery room is reported . Accumulation of atracurium in the intravenous line led to recurarization after flushing the line in the recovery room . A respiratory arrest with severe desaturation and bradycardia occurred . Circumstances leading to this event and the mechanisms enabling a <D>neuromuscular blockade</D> to occur , following the administration of a small dose of relaxant , are discussed .
\\\\
\textcolor{teal}{\textbf{[Example 2]}} \\
\textbf{The given sentence:} \\
\{context\}
\\
\textbf{The labeled sentence: }
    \end{tabular}
    \caption{The prompt for LLM-revision on BC5CDR.}
    \label{tab:baseline_prompt_bc5}
\end{table*}